\documentclass[runningheads]{llncs}

\usepackage{eccv}

\makeatletter
\newcommand{\printfnsymbol}[1]{%
  \textsuperscript{\@fnsymbol{#1}}%
}
\makeatother

\usepackage{eccvabbrv}
\usepackage{enumitem}

\usepackage{graphicx}
\usepackage{booktabs}

\usepackage[accsupp]{axessibility}  
\def\*#1{\mathbf{#1}}

\usepackage{svg}
\usepackage{multirow}
\usepackage{adjustbox}
\usepackage{wrapfig}
\usepackage{bmpsize}

\usepackage{hyperref}
\usepackage{orcidlink}

\begin{document}

\title{Can Your Generative Model Detect Out-of-Distribution Covariate Shift?} 

\titlerunning{CovariateFlow}

\author{Christiaan Viviers\inst{1}\orcidlink{0000-0001-6455-0288} \and
Amaan Valiuddin\inst{1}\thanks{Equal contribution}\orcidlink{0009-0005-2856-5841} \and
Francisco Caetano\inst{1}\printfnsymbol{1}\orcidlink{0000-0002-6069-6084} \and 
Lemar Abdi\inst{1}\and 
Lena Filatova\inst{2}\and
Peter de With\inst{1}\orcidlink{0000-0002-7639-7716} \and
Fons van der Sommen\inst{1}\orcidlink{0000-0002-3593-2356}}

\authorrunning{C.~Viviers \etal}

\institute{Eindhoven University of Technology, Eindhoven, The Netherlands
\email{c.g.a.viviers@tue.nl}
\and
Philips IGTs, Best, The Netherlands\\
}
\maketitle

\begin{abstract}


Detecting Out-of-Distribution~(OOD) sensory data and covariate distribution shift aims to identify new test examples with different high-level image statistics to the captured, normal and In-Distribution~(ID) set. Existing OOD detection literature largely focuses on \emph{semantic} shift with little-to-no consensus over covariate shift. Generative models capture the ID data in an unsupervised manner, enabling them to effectively identify samples that deviate significantly from this learned distribution, irrespective of the downstream task. In this work, we elucidate the ability of generative models to detect and quantify domain-specific covariate shift through extensive analyses that involves a variety of models. To this end, we conjecture that it is sufficient to detect most occurring sensory faults (anomalies and deviations in global signals statistics) by solely modeling high-frequency signal-dependent and independent details. We propose a novel method, CovariateFlow, for OOD detection, specifically tailored to covariate heteroscedastic high-frequency image-components using conditional Normalizing Flows~(cNFs). Our results on CIFAR10 vs. CIFAR10-C and ImageNet200 vs. ImageNet200-C demonstrate the effectiveness of the method by accurately detecting OOD covariate shift. This work contributes to enhancing the fidelity of imaging systems and aiding machine learning models in OOD detection in the presence of covariate shift. The code for CovariateFlow is available at \href{https://github.com/cviviers/CovariateFlow}{https://github.com/cviviers/CovariateFlow}.
\end{abstract}

\section{Introduction}
\label{sec:intro}

Identifying abnormal image statistics is critical for deploying precise sensing technology and reliable machine learning. Out-of-Distribution~(OOD) detection methods model the available data or a set of In-Distribution~(ID) features, to identify test examples drawn from a different distribution. Notably, generative models offer an unsupervised paradigm to model without making explicit assumptions on the form of the OOD data. With a plethora of possible covariates (abnormal variations in high-level image statistics) and potential downstream machine learning image applications, unsupervised generative modelling is a promising approach for general OOD detection. The prevailing approaches for OOD detection predominantly focus on the semantic contents of the image data. Therefore, this study elucidates covariate shifts, \emph{i.e.} the change in distribution of high-level image statistics (covariates) subject to consistent low-level semantics.

\begin{figure}[t]
\centering
\begin{subfigure}{.6\textwidth}
  \includegraphics[width=7.3cm]{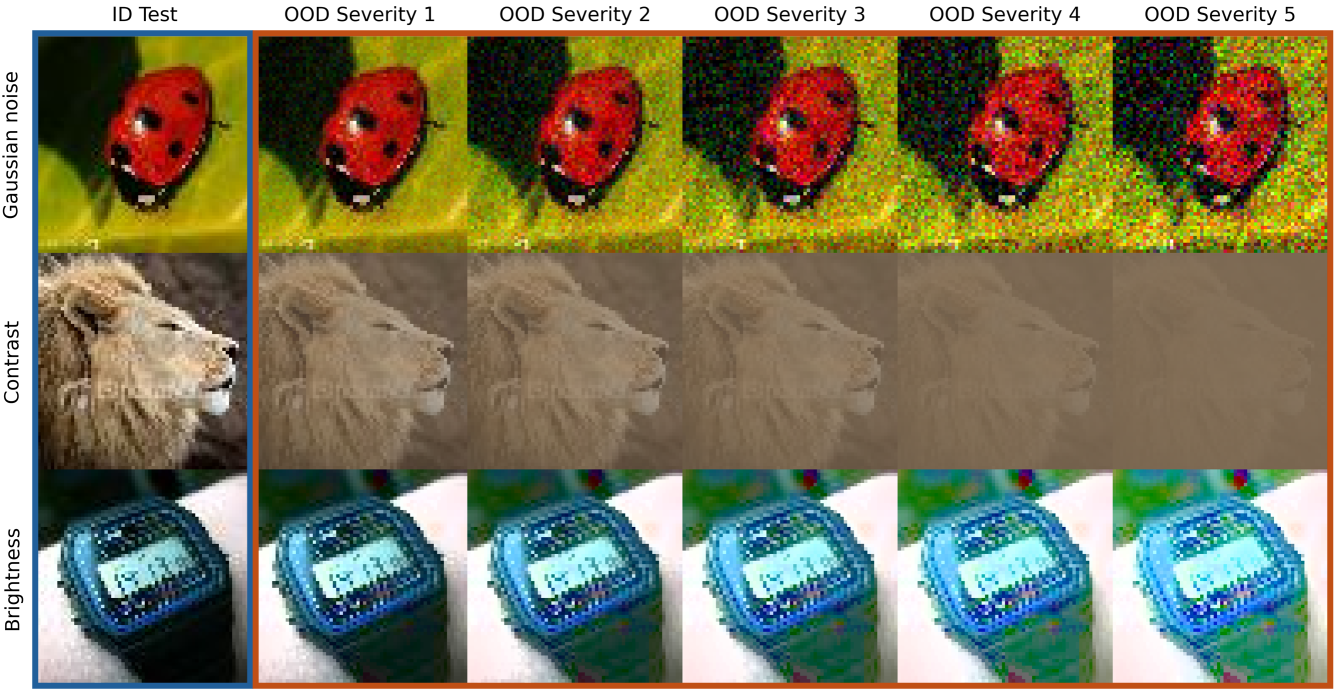}
  \caption{ImageNet200(-C) at 5 severity levels}
  \label{fig:imagenet_c}
\end{subfigure}%
\hspace{0.1cm}
\begin{subfigure}{.35\textwidth}
  \centering
  \includegraphics[width=4.5cm]{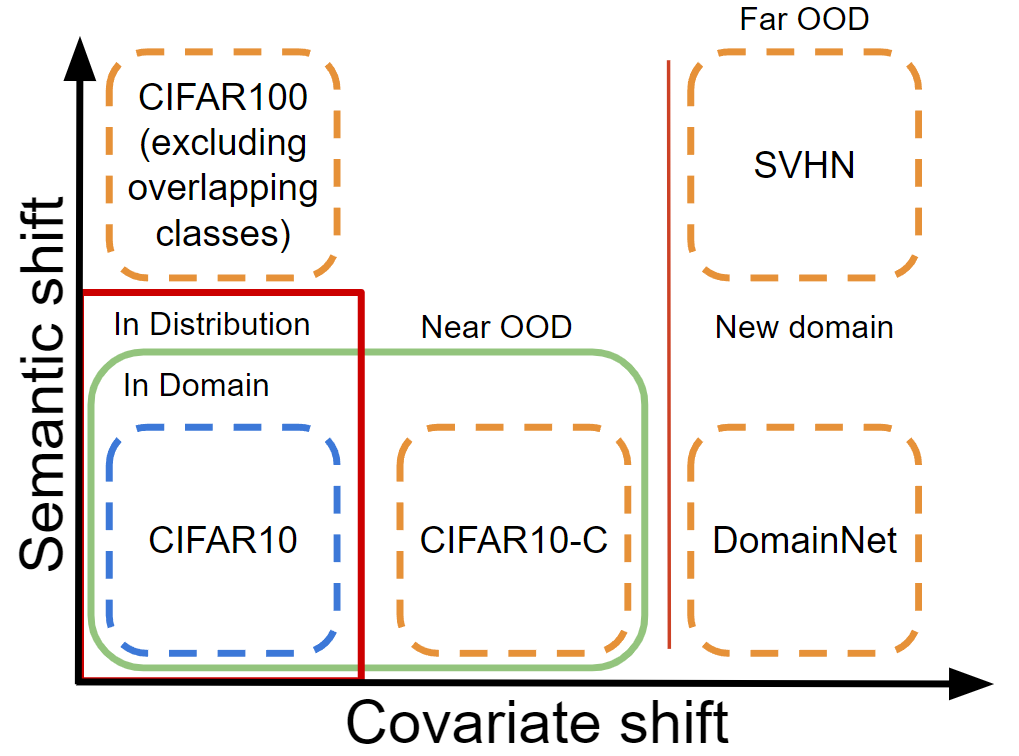}
  \caption{In-domain covariate shift (Near OOD) vs. covariate shift across domains (Far OOD).}
  \label{fig:covariate_shift}
\end{subfigure}
\caption{Illustrations of in-domain covariate shifts. (a) ID ImageNet200 and different degradations and severity levels of ImageNet200-C as OOD. (b) Covariate and semantic shift in terms of popular computer vision datasets.}
\label{fig:covariate_shift_images}
\end{figure}
Likelihood-based methods, such as Normalizing Flows~(NFs), offer an intuitive way of OOD detection by evaluating the likelihood of test samples. However, as evidenced in previous research~\cite{normalizingfail}, NFs have exhibited limitations in effective OOD detection, often assigning higher likelihoods to OOD samples. Various works have explored this phenomenon and proposed alternative methods to direct likelihood estimation~\cite{detectingusingtypicality, your_classifier}. Recent theoretical investigations~\cite{zhang2021understandingfailuresOOD} indicate that these methodologies are inherently susceptible to certain types of OOD data. Moreover, the metrics employed for evaluation exhibit a predisposition towards specific categories of OOD data, suggesting an intrinsic limitation in the current approach to OOD detection. In this study we explore this phenomena while improving the covariate OOD detection capabilities of NFs. Additionally, we address this shortcoming by proposing to unify the log-likelihood~(LL)-based metric with the \emph{typicality} score~\cite{chali2023typicality} in a simple Normalized Score Distance~(NSD). Other generative models have in various contexts been applied to the task of semantic OOD detection, ranging from density-based methods~\cite{kingma2022autoencoding, dinh2016density, yang2021generalized} to different reconstruction-based models~\cite{wyatt2022anoddpm}. However, OOD covariate shift within the context of generative models remains largely unexplored.

We indicate two branches of covariate shift: (1)~domain covariate shift, such as images in different styles (\emph{e.g.} natural vs. sketch) and (2)~domain-specific covariate shift (also known sensory anomalies~\cite{yang2021generalized}), images under different sensory conditions (\emph{e.g} different lighting, cameras or sensory level degradation (Figure~\ref{fig:covariate_shift_images}). Covariate shifts are recognized for their potential to significantly degrade the predictive performance of the model, where in some specialized imaging applications it can indicate system failure. Detecting these covariate factors and the distribution shifts under consistent semantic content~\cite{tian2021exploringcovariate}, will enhance the safety and reliability of imaging systems in diverse fields and the machine learning systems built on-top of these images~\cite{adhikarla2023robust, MLfails, wang2023adversarial}. This necessitates their detection, and if possible, the quantification of its severity. To this end, we are, to the best of our knowledge, the first to implement unsupervised, domain-specific OOD covariate shift detection.

Images across different applications can demonstrate complex noise patterns and variability due to factors such as equipment variations, environmental conditions, and the specific nature of the imaged objects or scenes~\cite{goyal2018noise_issues}. A novel and effective strategy for improving OOD detection should utilize the data-dependent (heteroscedastic) noise that is present in the signal. This inherent noise serves as a rich source of information that can be exploited to differentiate between ID and OOD samples. In fact, the noise patterns in images can encode subtle differences that may not be apparent from the semantic content in the image alone.

To address these challenges and leverage the nuanced information encoded in noise patterns across various imaging applications, we propose a streamlined approach that models the conditional distribution between low-frequency and high-frequency signal components. This method contrasts with conventional techniques that attempt to model the entire signal distribution, which may inadvertently obscure critical covariate details. We employ a simple filtering approach that segregates the image into distinct low-frequency and high-frequency components. By focusing on the interaction between these frequency components, the proposed approach effectively detect covariate shifts. The research contributions of this work are as follows.

\begin{itemize}[label=\textbullet]
   \setlength\itemsep{0em}
   \item \emph{Unsupervised OOD covariate shift} detection with comprehensive evaluation of generative model using CIFAR10(-C) and ImageNet200(-C) 
   \item \emph{CovariateFlow}: A novel application of conditional Normalizing Flows for high-frequency heteroscedastic image-component density modelling.
   \item \emph{Normalized Score Distance}: which unifies Typicality and Log-likelihood as a general metric for OOD detection in normalizing flows.
   \item \emph{Accurate detection of distribution shift} and sensory-level changes indicating high sensitivity for potential fault detection. 
\end{itemize}

\section{ Background and Related Work}
\label{sec:related}

\subsection{Semantic Out-of-Distribution Detection}\label{sec:ood}
Approaches to OOD detection are generally divided into two categories: \emph{supervised}, which necessitates labels or OOD data, and \emph{unsupervised}, which relies solely on in-distribution data. Although semantic OOD detection does not constitute the core focus of this study, we nevertheless provide a concise overview of the recent developments, since these methodologies hold the potential to translate to covariate OOD detection. For an in-depth exploration of OOD detection methodologies, we refer to the comprehensive review by Yang~\etal~\cite{yang2021generalized}.

\subsubsection{Explicit Density Methods:} A straightforward method for OOD detection involves the use of a generative model, $p(\*x; \theta)$ parameterized by $\theta$ and trained to fit a given distribution over data $\*x$. The process evaluates the likelihood of new, unseen samples under this model with the underlying assumption that OOD samples will exhibit lower likelihoods compared to those that are ID. The Evidence Lower Bound~(ELBO) employed in Variational Auto Encoders~(VAEs)~\cite{kingma2022autoencoding} can be used for OOD detection by evaluating a lower bound on the likelihood of a test sample. Plumerault~\etal~\cite{plumerault2021avae} introduced the Adversarial VAE -- a novel approach that marries the properties of VAEs with the image generation quality of Generative Adversarial Networks~(GANs), thereby offering a robust auto-encoding model that synthesizes images of comparable quality to GANs while retaining the advantageous characteristics of VAEs. 

Unlike VAEs, Normalizing Flows~(NFs)~\cite{kobyzev2020normalizing} offer exact and fully tractable likelihood computations. With the introduction of coupling layers~\cite{dinh2016density}, NFs can be arbitrarily conditioned and seem to be excellent contenders for conditional OOD detection. However, as evidenced in previous research~\cite{nalisnick2018deep}, NFs have exhibited limitations in effective OOD detection, often assigning higher likelihoods to OOD samples. This limitation has been associated with an inherent bias in the flow model architectures, which tends to prioritize modeling local pixel correlations over the semantic content of the data~\cite{normalizingfail}. Exploration by Gratwohl \textit{et al.}~\cite{your_classifier} and Nalisnick \textit{et al.}~\cite{detectingusingtypicality} posits that this phenomenon can be attributed to the fact that ID images are not high-likelihood samples but, rather, constituents of the \textit{typical set} of the data distribution. Consequently, the investigation into methods that assess the \textit{typicality}~\cite{chali2023typicality} of data instances, as an alternative to direct likelihood estimation, has gained traction. Despite empirical evidence demonstrating the efficacy of typicality in OOD benchmarks~\cite{chali2023typicality}, recent theoretical investigations~\cite{zhang2021understandingfailuresOOD} indicate that these methodologies have inherent susceptibilities to specific OOD types and an evaluative bias towards particular OOD categories, thereby underscoring the complexity of OOD detection.

\subsubsection{Image Reconstruction-based Methods:} These OOD detection methods are based on the principle that models are less effective at accurately reconstructing data that significantly deviates from the training distribution. Graham~\etal~\cite{graham2023denoising} improves on an innovative approach to OOD detection that leverages the potent generative prowess of recent denoising diffusion probabilistic models~(DDPMs)~\cite{ho2020denoising,nichol2021improved}. Unlike prior reconstruction-based OOD detection techniques that necessitated meticulous calibration of the model's information bottleneck~\cite{denouden2018improving, pimentel2014review,zong2018deep}, their method utilizes DDPMs to reconstruct inputs subjected to varying degrees of noise. In this work, we implement this DDPM method as baseline for OOD covariate shift detection.\newline

\subsection{Covariate shift}\label{sec:cov_shift}

In essence, covariate shift refers to the phenomenon where images share consistent semantic content (\emph{i.e.} similar subjects), and yet, are captured under varying imaging conditions. The degree of variation in these conditions signifies the magnitude of the shift. For example, a minor shift might involve images of a subject under varying lighting conditions, while a more substantial shift, such as transitioning from natural images to graphical sketches of the same subject, exemplifies a transition towards domain shift (Figure~\ref{fig:covariate_shift}). This paper concentrates on in-domain covariate shifts, as these scenarios represent instances where machine learning silently fails~\cite{david2010impossibility, adhikarla2023robust, MLfails, wang2023adversarial}. 

\indent In related work on covariate shift detection, Averly~\etal.~\cite{averly2023unified} adopt a model-centric approach to address both covariate and semantic shifts, suggesting a methodology for identifying instances that a deployed machine learning model, such as an image classifier, fails to accurately predict. This strategy implies that the decision to detect, and potentially exclude, a test example, is dependent on the specific model in question. While being effective for well-established machine learning models, this method inherently links the detection of shifts to the peculiarities of the individual model, resulting in each model having its unique set of criteria for rejecting data, which may vary broadly, even when applied to the same dataset. A significant drawback of this approach is its reliance on a robust pre-trained model, which poses a challenge for scenarios where identifying covariate shift is the primary objective, leaving such cases without a viable solution.\newline
\indent Generalized ODIN~\cite{hsu2020generalized_odin} is another direction of work that adopts the model-centric approach. This method replaces the standard classification head and, instead, decomposes the output into scores to behave like the conditional probabilities for the semantic shift distribution and the covariate shift distribution. This approach is then only evaluated on out-of-domain covariate shift such as the DomainNet~\cite{domainnet} benchmark. Follow-up work by Tian~\etal~\cite{tian2021exploringcovariate} further explores calibrating the confidence functions proposed in \cite{hsu2020generalized_odin}, which realizes improvements on both semantic and covariate OOD detection. They additionally apply their refinement on in-domain covariate shift, such as CIFAR10 vs. CIFAR10-C.\newline
\indent Besides the above-mentioned work, covariate shift has been studied predominantly from a robustness perspective~\cite{adhikarla2023robust} or in a domain adaptation setting~\cite{gretton2008covariate, ganin2016domainadversarial}. The defense against adversarial attacks~\cite{wang2023adversarial} is another research direction that falls in the domain of covariate shift. This perspective stems from the recognition that adversarial examples, by nature, often represent data points that deviate significantly from the distribution observed during model training, thereby inducing a form of covariate shift. Researchers have leveraged insights from adversarial robustness~\cite{wang2023adversarial} to devise methods that can identify and mitigate the effects of such shifts, focusing on enhancing model reliability and security against deliberately crafted inputs designed to deceive. Fortunately, the shift introduced is completely artificial and typically a shift targeted at a specific model.

\subsection{Normalizing Flows~(NFs)}

Consider an image sampled from its intractable distribution as $\mathbf{x}$$\,\sim\,$$P_\mathbf{X}$. Additionally, let us introduce a simple, tractable distribution $p_\mathbf{Z}$ of latent variable $\mathbf{z}$ (which is usually Gaussian). Normalizing Flows utilize $k$ consecutive bijective transformations $f_k:\mathbb{R}^D\rightarrow\mathbb{R}^D$ as $\*f = f_K \circ \ldots \circ f_{k} \circ \ldots \circ f_1$, to express exact log-likelihoods 
\begin{equation}\label{eq:nf}
    \log p(\mathbf{x}) = \log p_{\*Z}(\mathbf{z}_0) - \sum_{k=1}^K \log \left| \det \frac{\mathrm{d} f_k(\mathbf{z}_{k-1})}{\mathrm{d} \mathbf{z}_{k-1}} \right|,
\end{equation}
where $\*z_k$ and $\*z_{k-1}$ are intermediate variables, and $\mathbf{z}_0 = \*f^{-1}(\*x)$.

Numerous bijections have been introduced which balance expressivity and have a simple evaluation of the Jacobian determinant in Equation~(\ref{eq:nf}). Specifically, coupling flows have seen much success~\cite{dinh2016density,ardizzone2019guided}. Since they can be parameterized through arbitrary complex functions, we explore conditioning the flow on the frequency components of an image.

\subsection{Typicality} \label{sec:typicality}

Examining sequences of $N$ independent and identically distributed (i.i.d.) datapoints $\mathbf{x}_n$, the \emph{typical set} comprises all $\mathbf{x}_n$ that satisfy
\begin{equation}\label{eq:typical}
H(\mathbf{X})-\epsilon \leq -\frac{1}{N} \sum_{n=1}^N \log_2 p(\mathbf{x}_n) \leq H(\mathbf{X})+\epsilon,
\end{equation}
where $\epsilon$ represents an arbitrarily small value and $H(\mathbf{X})$ denotes the Shannon entropy of the dataset. In other words, the empirical entropy of the set approaches to the entropy rate of the source distribution. Leveraging the Asymptotic Equipartition Property~(AEP), it is deduced that
\begin{equation}
    \frac{1}{N} \sum_{n=1}^N \log_2 p(\mathbf{x}_n)\rightarrow  H(\mathbf{X}) \quad \text{s.t. } N\rightarrow\infty,
\end{equation}
leading to the conclusion that the probability of any sequence of i.i.d. samples of sufficient length approaches unity. Thus, despite the typical set representing merely a small subset of all potential sequences, a sequence drawn from i.i.d. samples of adequate length will almost certainly be considered typical~\cite{thomas2006elements}. 

In various studies, indications have emerged that NFs perform poorly when the likelihood is utilized as a metric for detecting OOD samples~\cite{caterini2022entropic,zhang2021understandingfailuresOOD,normalizingfail,nalisnick2018deep}. It can be argued that datasets are a typical sequence of samples, rather than high in likelihood, also known as the Typical Set Hypothesis~(TSH). Therefore, in the recent work by Nalisnick~\etal~\cite{detectingusingtypicality}, an innovative approach is proposed for OOD detection that leverages typicality as an evaluation metric in lieu of likelihood. This methodology was further refined in subsequent studies~\cite{your_classifier}, introducing \emph{approximate mass}. Motivated by the fact that typical samples are localized in high-mass areas on the PDF, the metric evaluates the gradient of the LL w.r.t the input data, also known as the \emph{score}. It can be expressed mathematically as \(\left\| {\partial L(\*x;\theta)}/{\partial \*x} \right\|\), where \(\*x\) denotes the input, \(L\) the evaluated LL by the model parameterized by \(\theta\), and \(\left\|.\right\|\) represents the Euclidean norm. Despite some criticism on TSH~\cite{zhang2021understandingfailuresOOD}, this metric demonstrates superior performance in OOD detection across various benchmarks~\cite{chali2023typicality,your_classifier}.

\section{Approach}
\label{sec:approach}
\subsection{Definition of Covariate Shift}
Formally, semantic- and in-domain covariate shifts can be delineated as follows. Consider samples from the training distribution, $\*x \sim P_{\*X}$, and anomalous data from an OOD source $\hat{\*x} \sim P_{\hat{\*X}}$, subject to a low-pass filter ($l$, $l:\mathbb{R}\rightarrow\mathbb{R}$) to obtain the low-frequency components, $\*x_\text{L} = l(\*x)$ and the high-frequency components $\*x_\text{H} = \*x - l(\*x)$. Semantic shift is characterized by a discrepancy in the marginal probability distributions, $P_{\*X_{\text{L}}} \neq P_{\hat{\*X}_{\text{L}}}$, when the conditional probability distributions of high-frequency components remain consistent, $P_{\*X_{\text{H}}|\*X_{\text{L}}} \approx P_{\hat{\*X}_{\text{H}}| \hat{\*X}_{\text{L}}}$. Conversely, covariate shift is identified when the conditional probability distributions diverge, $P_{\*X_{\text{H}}|\*X_{\text{L}}} \neq P_{\hat{\*X}_{\text{H}}|\hat{\*X}_{\text{L}}}$, but the marginal probability distributions of the low-frequency components remain the same $P_{\*X_{\text{L}}}\approx P_{\hat{\*X}_{\text{L}}}$. Furthermore, these definitions hold with in the supervised setting with predefined targets ($\*Y$).

\subsection{CovariateFlow}\label{sec:covariate_flow}
In the development of methodologies for detecting covariate shift within datasets, several critical factors must be meticulously considered to ensure efficacy and accuracy. Firstly, (1)~the process of resizing images can significantly alter the distribution of high-frequency statistics, potentially obscuring key data characteristics. Secondly, (2)~the inherent nature of encoding architectures, which essentially function as low-pass filters~\cite{low_pass_encoder}, may limit their capacity to fully capture the complex distribution of noise present within the data. This limitation is particularly relevant as covariate shifts often manifest through alterations in the general image statistics, thereby necessitating a method capable of discerning such nuances. Thirdly, (3)~the utilization of \textit{only} log-likelihood-based evaluation in NFs, has proven to have a predisposition towards low-level semantics and is more sensitive to high-frequency statistics. An effective method should be sensitive to covariate shifts affecting all frequency bands, from noise degradations to contrast adjustments.

\begin{figure}[!htbp]
\centering
\begin{subfigure}{.58\textwidth}
  \centering
 \includegraphics[width=1.0\linewidth]{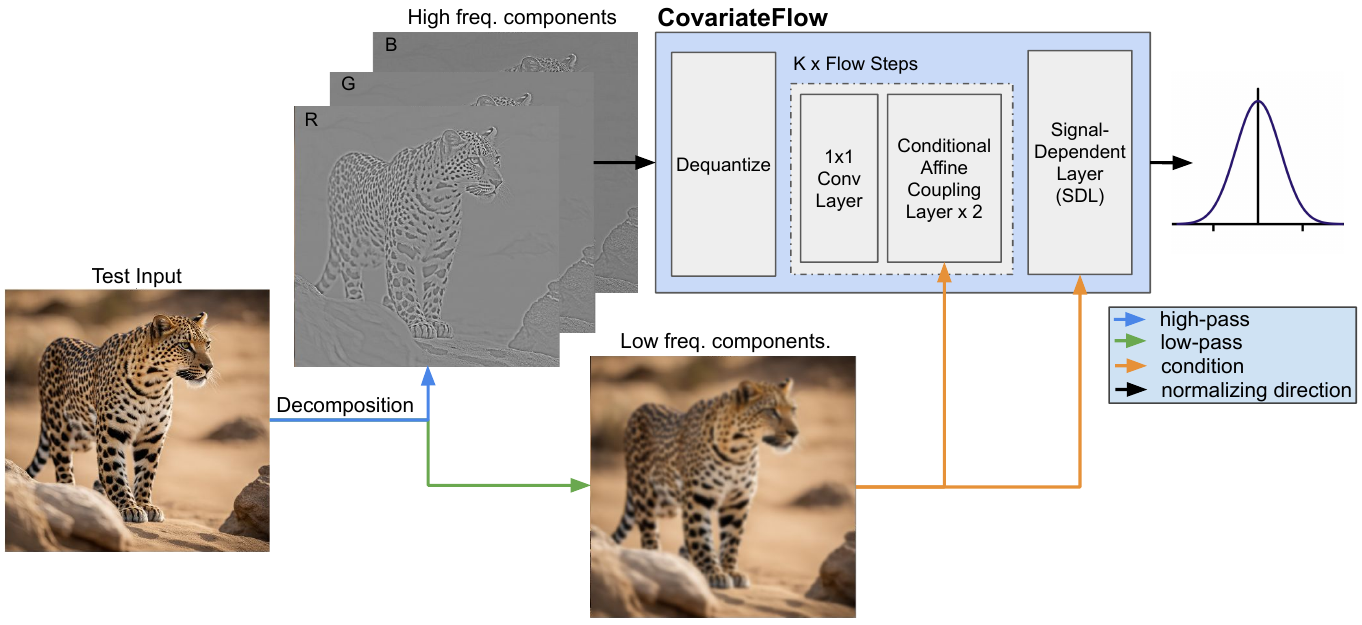}
  \caption{High-level model architecture
  \label{fig:arch}}
\end{subfigure}%
\hspace{0.25cm}
\begin{subfigure}{.38\textwidth}
  \centering
  \includegraphics[width=1.0\linewidth]{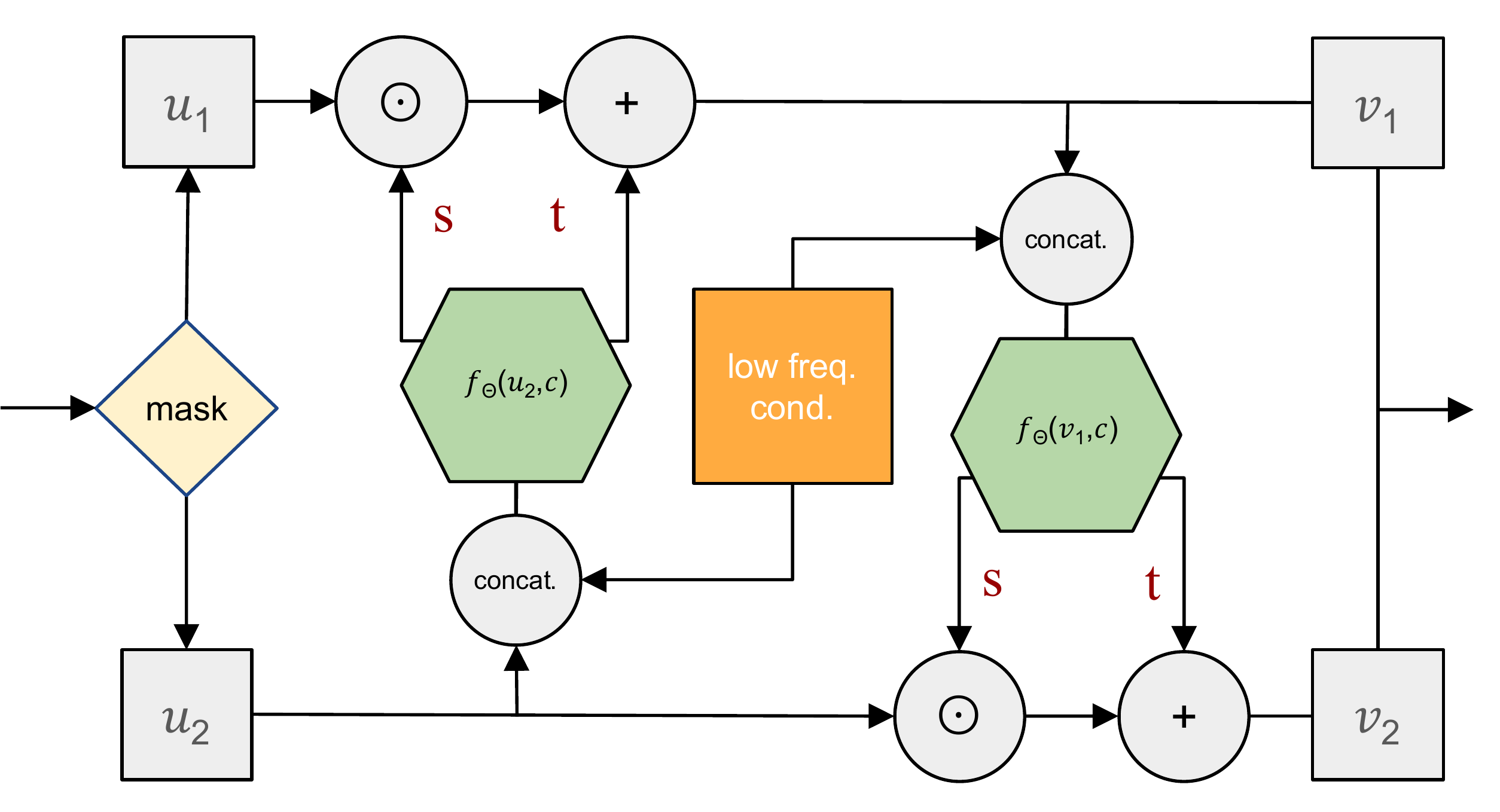}
  \caption{Conditional coupling layer (2 steps). $f_{\Theta}(\*x, \*c)$ details the neural network.\label{fig:coupling}}
\end{subfigure}
\caption{Illustration of CovariateFlow architecture and model components.}
\label{fig:covariateflow_arch}
\vspace{-0.2cm}
\end{figure}

\indent In light of the above considerations, Normalizing Flows~(NFs) emerge as a particularly suitable candidate for modeling the imaging features essential for detecting covariate shift. NFs are distinct in that they abstain from any form of down-sampling or encoding processes to preserve their bijective property. It is also recognized that NFs prioritize pixel correlations over semantic content~\cite{normalizingfail}. However, given the expectation that covariate shift involves changes in high-frequency image statistics, accurately modeling the complete image distribution—including both low-frequency semantics and high-frequency components—presents significant challenges, particularly given the relatively limited capacity of NFs compared to more recent generative models~\cite{ho2020denoising, nichol2021improved, song2021scorebased}.\newline
\indent To address these challenges, we introduce a novel method that simplifies the modeling of components critical for covariate shift detection. Our approach involves a filtering strategy that divides the image into separate low-frequency and high-frequency components, thereby allowing the detection system to concentrate specifically on the high-frequency elements to improve detection capabilities. Consider an input signal $\*x$ and the low-pass filter $l$, the conditional distribution of high-frequency components ($\mathbf{x}_\text{H}  = \*x - l(\*x)$) given the low-frequency components ($\mathbf{x}_\text{L} = l(\*x)$) . By recognizing that certain high-frequency components are correlated with low-frequency signals, we can model this relationship conditionally. On this premise, we develop CovariateFlow (Figure~\ref{fig:arch}), a novel approach of modeling the conditional distribution between high-frequency and low-frequency components using conditional NFs as
\begin{equation}\label{eq:nf_freq}
    \log p(\mathbf{x}_\text{H} \vert \mathbf{x}_\text{L} ) = \log p_{\*Z}(\mathbf{z}_0) - \sum_{k=1}^K \log \left| \det \frac{\mathrm{d} f_k(\mathbf{z}_{k-1}, \*x_L)}{\mathrm{d} \mathbf{z}_{k-1}} \right|.
\end{equation}
This formulation sets the foundation for a detection system that is finely attuned to the nuances of covariate shift, enhancing its ability to identify and respond to shifts in high-frequency image statistics. The proposed model is predominantly defined by (1)~a signal-dependent layer~(SDL)~\cite{noiseflow}, (2)~conditional coupling flow~\cite{ardizzone2019guided}, (3)~an unconditional 1$\times$1 convolutional (conv.) layer~\cite{kingma2018glow} and (4)~uniform dequantization,. The SDL layer and conditional coupling layer are additionally conditioned on $\*x_\text{L}$. The 1$\times$1-convolution and conditional coupling flow is repeated $K$ times depending on the dataset at hand. We employ a Gated ResNet~\cite{ho2019flow++} as $f_\Theta$ and a checkerboard masking strategy~\cite{dinh2016density} in our coupling layers. Figure~\ref{fig:arch} depicts a high-level overview of the model architecture. We employ a simple Gaussian filter for $l$, to decompose the signal into low-frequency and high-frequency components. To minimize any assumptions about the high-frequency components, we use a conventional Gaussian kernel. A kernel with a standard deviation ($\sigma$) of one proved to yield the best performance. The coupling layers are depicted in Figure~\ref{fig:coupling} and the model totals a mere 945,882 trainable parameters with 16 coupling layers. For a detailed description on training details, ablation experiments and inference time comparisons, we refer to Section~\ref{sec:supp_ablation} in the supplementary material. The code for the model will be made publicly available.
\subsection{Unifying Log-likelihood and Typicality}\label{sec:unify}

The inductive bias of NFs towards structural complexity when evaluating with LL has been discussed in Section~\ref{sec:typicality}. As an alternative, evaluation on typicality using the gradient of the LL w.r.t. the input data, has shown improvements in semantic OOD detection over LL~\cite{chali2023typicality, grathwohl2019your}. However, it is understood that the metric and model are similarly biased towards certain categories of data~\cite{zhang2021understandingfailuresOOD}. As such, we propose to combine LL evaluation with the Typicality score to overcome the limitations of each approach individually. Our approach standardizes both the LL and the typicality scores in terms of their respective training statistics. After standardization, we can transform each metric into an absolute distance from the expected mean. The LL distance and gradient score distance can then simply be added to obtain a unified distance. In this manner, the evaluation is sensitive to all deviations, rather than only being lower in score, thereby reducing the effect of the biases of the respective metrics. The following paragraph gives the details of the mentioned approach.

Consider a sample $\*x\sim P_\*X$ with log-likelihood $\log p(\*x)$. Furthermore, we denote the magnitude of the gradients as \(||\nabla_\*x \log p(\*x)||\), i.e. the approximate score. The means for the empirical likelihoods are determined through $\mu_{L}=\mathbb{E}_{P_\*X}[\log p(\*x)]$, and of the approximate scores with $\mu_{T}=\mathbb{E}_{P_\*X}[\,||\nabla_\*x \log p(\*x)||\,]$. Similarly, we can denote $\sigma^2_L=\mathbb{E}_{P_\*X}[(\*x-\mu_L)^2]$ and $\sigma^2_T=\mathbb{E}_{P_\*X}[(\*x-\mu_T)^2]$ for their respective variances. We can then obtain the Normalized Score Distance~(NSD) for a new sample $\*x^*$ as the summation of the standardized L1-norms through
\begin{equation}\label{eq:nsd}
\operatorname{NSD}(\*x^*) = \left| \frac{\log p(\*x^*) - \mu_L}{\sigma_L} \right| + \left| \frac{||\nabla_\*x \log p(\*x^*)|| - \mu_T}{\sigma_T} \right|.
\end{equation}
Figure~\ref{fig:nsd_result} depicts this procedure for two degradation types on CIFAR10. 

\subsection{Datasets}

\textbf{CIFAR10(-C) \& ImageNet200(-C):} CIFAR10~\cite{cifar10} and ImageNet200 with their respective corrupted counterparts, CIFAR10-C~\cite{cifar10c} and ImageNet200-C~\cite{cifar10c}, serve as exemplary datasets for developing and evaluating unsupervised covariate shift detection algorithms. CIFAR10 and ImageNet200 provide a collection of images that encompass a broad range of in-distribution covariate shifts, ensuring a suitable level of diversity. On the other hand, the corrupted versions introduce real-world-like (undesired) degradations, such as noise, blur, weather, and digital effects. Figure~\ref{fig:imagenet_c} depicts 3 of the 15 effects employed in the ImageNet200(-C) dataset. Images are utilized in their original resolution at 64~$\times$~64~pixels.  CIFAR10-C consists of 19 corruptions in total with images at 32~$\times$~32~pixels. This setup enables testing the covariate shift detecting performance across multiple distortion types and severity levels. In all our experiments we train the models only on the original dataset's training set and then test it against \textit{all} of the corruptions at every severity level. For CIFAR10 this is the original dataset's test set (ID test) and CIFAR10-C's 19 corruptions at 5 severity levels (95 OOD test sets). Similarly, we treat the ImageNet200 test set as ID test and the 15 corruptions at 5 severity levels from ImageNet200-C as 75 OOD datasets. The datasets follow the OpenOOD~\cite{zhang2023openood} benchmarks\footnote{\href{https://github.com/Jingkang50/OpenOOD}{https://github.com/Jingkang50/OpenOOD}}.

\section{Experiments }
\label{sec:experiments}
The following section describes the conducted experiments and presents the key results obtained in our investigation. Further detailed experimental results can be found in the supplementary materials. Specifically, we present results on CIFAR10 (Section~\ref{sec:supp_detailed_results}), ImageNet200 (Section~\ref{sec:supp_detailed_results_tin}) and extensive ablation experiments with the proposed CovariateFlow (Section~\ref{sec:supp_ablation}).

\subsection{Evaluation Metrics \& Models}

To evaluate the model's ability to detect OOD covariate shifts, we utilize metrics commonly found in related work: the Area Under the Receiver Operating Characteristic~(AUROC) curve and the False Positive Rate~(FPR) at a 95\% True Positive Rate~(TPR). In all our experiments with CIFAR10(-C) and ImageNet200(-C), we use the designated test set (10k samples) to compute each metric. Our contributions include contextualizing the VAE, AVAE, GLOW evaluated with log-likelihood and the DDPM with the reconstruction loss, within OOD covariate shift as baseline models. Furthermore, we evaluate GLOW using typicality and the proposed NSD and CovariateFlow with all the aforementioned metrics. Most models are trained from scratch on the ID data. For the VAE-FRL~\cite{cai2023out}, a method leading in semantic OOD detection, the available pretrained CIFAR10 weights\footnote{\href{https://github.com/mu-cai/FRL}{https://github.com/mu-cai/FRL}} are employed. A detailed description of the implemented models can be found in Section~\ref{sec:supp_implemntation_details} of the supplementary materials.

\subsection{Covariate Shift in CIFAR10 and ImageNet200}
Table~\ref{tab:auroc_per_level} showcases various models and their averaged AUROC across all the degradations per CIFAR10-C/ImageNet-C severity level. While some models excel in handling specific types of degradation, only the overall performance is truly relevant, as one typically cannot predict the type of perturbation that will occur in real-world settings. A detailed breakdown of the results per perturbation is shown in Section~\ref{sec:supp_detailed_results} of the supplementary materials. 
\begin{figure}[tbp]
\centering
  \includegraphics[width=12cm]{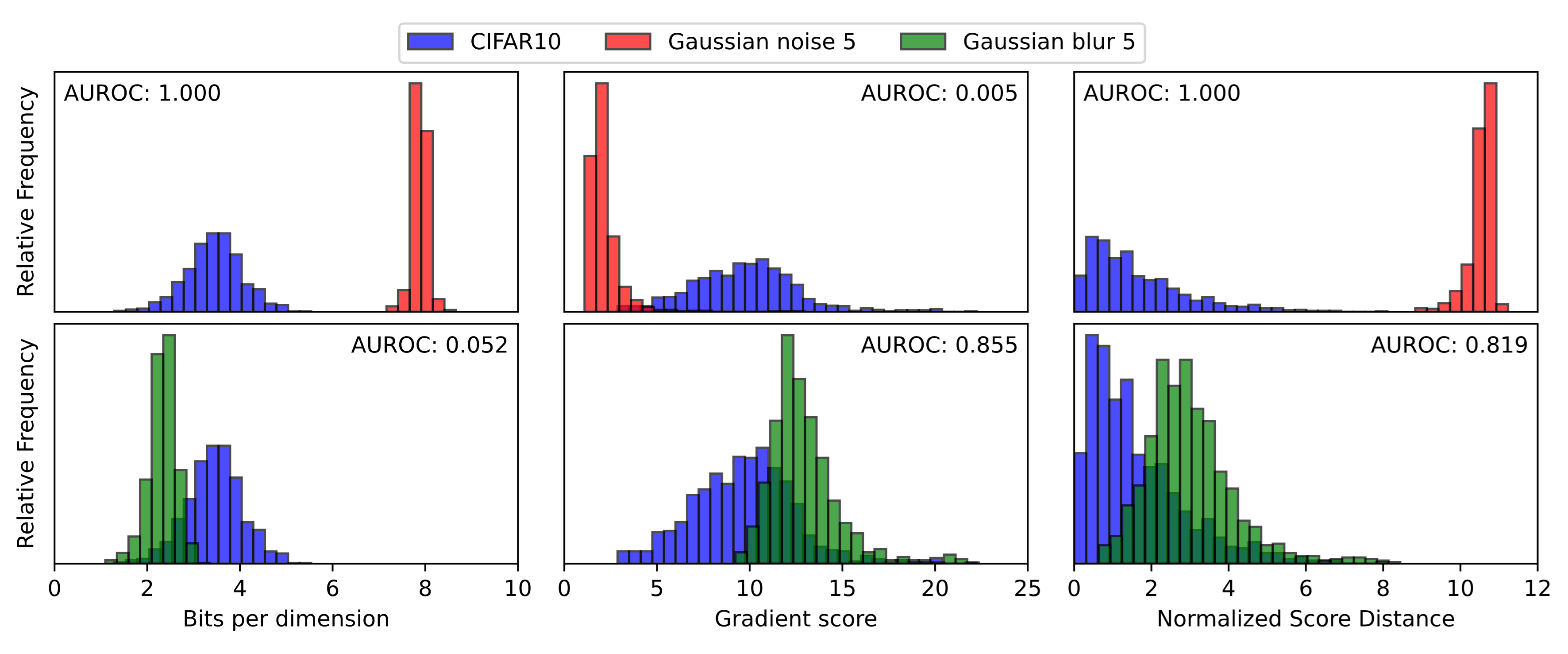}
  \caption{Results obtained with GLOW evaluated with Log-likelihood, Gradient score and NSD on CIFAR10 (test set) vs. CIFAR10-C (test set). Top row: \textbf{Gaussian Noise 5}, Bottom row:  \textbf{Gaussian Blur 5}.}\label{fig:nsd_result}
\end{figure}
\begin{table}[!bp]
\centering
{\resizebox{1.0\linewidth}{!}{
\begin{tabular}{clcccccc} 
\toprule
& & \multicolumn{5}{c}{\underline{CIFAR10-C OOD Severity Levels}} & Average\\
&Models &\enspace1\enspace&\enspace2\enspace&\enspace3\enspace&\enspace4\enspace&\enspace5\enspace& AUROC$\uparrow$\slash FPR95$\downarrow$\\
\midrule
\parbox[t]{5mm}{\multirow{13}{*}{\rotatebox[origin=c]{90}{CIFAR10 ID}}} & \textbf{Reconstruction}\\
\midrule
&DDPM~\cite{graham2023denoising} (T150: LPIPS)&  55.1 & 59.9 & 63.6 & 66.5 & 70.5 & 63.1\slash83.9\\
&DDPM~\cite{graham2023denoising} (T20: LPIPS + MSE)& 58.2 & 63.8 & 69.0 & 71.0 & 75.6 & 67.5\slash75.2\\
\cmidrule{2-8}
&\textbf{Explicit Density}\\
\cmidrule{2-8}
&Vanilla VAE~\cite{kingma2022autoencoding} (SSIM + KL Div) & \enspace48.3\enspace&\enspace47.8\enspace&\enspace48.8\enspace&\enspace 50.3\enspace&\enspace49.5\enspace& 48.9\slash83.3\\
&AVAE~\cite{plumerault2021avae} (MSE + KL Div + Adv Loss) & \enspace 53.6\enspace &\enspace 58.0\enspace&\enspace60.2\enspace&\enspace63.9\enspace &\enspace 65.2\enspace& 60.2\slash73.1\\
&VAE-FRL~\cite{cai2023out} & 51.0 & 56.4 & 55.8 & 59.3 & 63.6 & 57.2/76.3\\
&GLOW~\cite{kingma2018glow} (LL) & 60.7 & 57.5 & 58.4 & 58.7 & 57.7 & 57.7\slash 69.5 \\
&GLOW~\cite{chali2023typicality} (Typicality) & 41.9 & 42.9 & 41.2 &  40.7 & 41.2 & 41.6\slash85.8 \\
&GLOW (NSD) &63.1 & 67.7 &68.9 & 70.9 & 75.6&  69.3\slash65.7\\
&CovariateFlow (LL) & 59.8 & 56.6 & 57.3& 58.5 &59.1& 58.3\slash63.5\\
&CovariateFlow (Typicality)  & 44.5 & 46.1 & 46.1&45.1&45.7 & 45.5\slash83.8\\
&CovariateFlow (NSD)  &\enspace\textbf{65.9}\enspace&\enspace\textbf{72.9}\enspace&\enspace\textbf{75.5}\enspace&\enspace \textbf{78.6}\enspace&\enspace\textbf{81.7}\enspace &\textbf{74.9}\slash\textbf{61.7}\\
\midrule 
& & \multicolumn{5}{c}{\underline{ImageNet200-C OOD Severity Levels}} & Average\\
&Models&\enspace1\enspace&\enspace2\enspace&\enspace3\enspace&\enspace4\enspace&\enspace5\enspace& AUROC$\uparrow$\slash FPR95$\downarrow$\\
\toprule
\parbox[t]{5mm}{\multirow{13}{*}{\rotatebox[origin=c]{90}{ImageNet200 ID}}} & \textbf{Reconstruction}\\
\cmidrule{2-8}
&DDPM~\cite{graham2023denoising} (T20: LPIPS + MSE)&\enspace48.6\enspace&\enspace56.9\enspace&\enspace65.1\enspace&69.7\enspace&\enspace74.0\enspace&62.9/75.8\\
\cmidrule{2-8}
&\textbf{Explicit Density}\\
\cmidrule{2-8}
&Vanilla VAE~\cite{kingma2022autoencoding} (SSIM + KL Div) &\enspace31.5\enspace&\enspace36.1\enspace&\enspace40.2\enspace&\enspace42.6\enspace&\enspace45.7\enspace&39.3/92.9\\
&AVAE~\cite{plumerault2021avae} (MSE + KL Div + Adv Loss) & 34.7 & 37.9 & 40.8 & 42.3 & 44.9 & 40.1\slash92.7\\
&GLOW~\cite{kingma2018glow} (LL) & 35.2 & 38.4 & 37.0 & 35.8 & 34.7 & 36.2\slash 81.7 \\
&GLOW~\cite{chali2023typicality} (Typicality) & 50.7 & 48.8 & 49.9 &  51.7 & 53.8 & 51.0\slash79.8 \\
&GLOW (NSD) &52.3 & 61.6 & 66.4 & 69.8 & 72.4 &  64.5\slash65.6\\
&CovariateFlow (LL) & 18.7 & 23.7 & 27.7 & 28.6 &29.0 & 25.5\slash86.9\\
&CovariateFlow (Typicality)  & \textbf{65.6} & 64.1 & 60.9 & 61.4 & 62.0 & 61.8\slash73.1\\
&CovariateFlow (NSD)  & 64.2 & \textbf{64.7} & \textbf{74.6}&\textbf{78.0}&\textbf{80.0}&\textbf{72.3}\slash\textbf{60.1}\\
\bottomrule 
\end{tabular}}}
\caption{Average AUROC scores of various methods on detecting the different severity levels of OOD covariate shift with the CIFAR10(-C) and ImageNet200(-C) dataset.}
\label{tab:auroc_per_level}
\end{table}

In Table~\ref{tab:auroc_per_level} it can be seen that models preserving the data dimension and maintaining the high-frequency signal components, such as the DDPM and NF-based approaches, perform best. ImageNet200-C contains fewer noise-based degradations than CIFAR10-C. The NF models evaluated with LL generally perform well on noise perturpowbations (Table~\ref{tab:glow_results_ll} and Table~\ref{tab:glow_results_ll_tin}) and because of this disparity in the types of degradations present in the datasets, LL evaluation exhibits a drop in average performance from CIFAR10 to ImageNet200. The VAE-FRL is designed to focus on semantic content and thus fails to accurately detect a change in general image statistics. It can be observed that CovariateFlow with NSD consistently outperforms the other methods at every severity level, realizing an average improvement of 5.6\% over GLOW on CIFAR10 and 7.8\% over GLOW on ImageNet200 when evaluated with NSD. This shows the strength of the proposed NSD metric, consistently improving over just LL or Typicality on both the GLOW model and CovariateFlow. Figure~\ref{fig:nsd_result} showcases an explicit example of how NSD consistently performs well under different degradations. 

Section~\ref{sec:supp_detailed_results} and Section~\ref{sec:supp_detailed_results_tin} presents a comprehensive evaluation of various methods for every OOD covariate shift type between the CIFAR10(-C) and ImageNet200(-C) datasets. Table~\ref{tab:aurocresults_cifar} focuses on the models' performances across three specific degradations (Gaussian Noise, Gaussian Blur, and Contrast) at five severity levels, that summarize the general results seen across all degradations. ImageNet200-C does not contain Gaussian Blur, but in general, the same trend can be observed between the two datasets for all the employed models. A complete comparison between all the models and their average performance per degradation type (averaged over severity levels) can be seen in Table~\ref{tab:models_per_degredation}.

\begin{figure}[!b]
\centering
\vspace{-0.5cm}
  \includegraphics[width=0.9\textwidth]{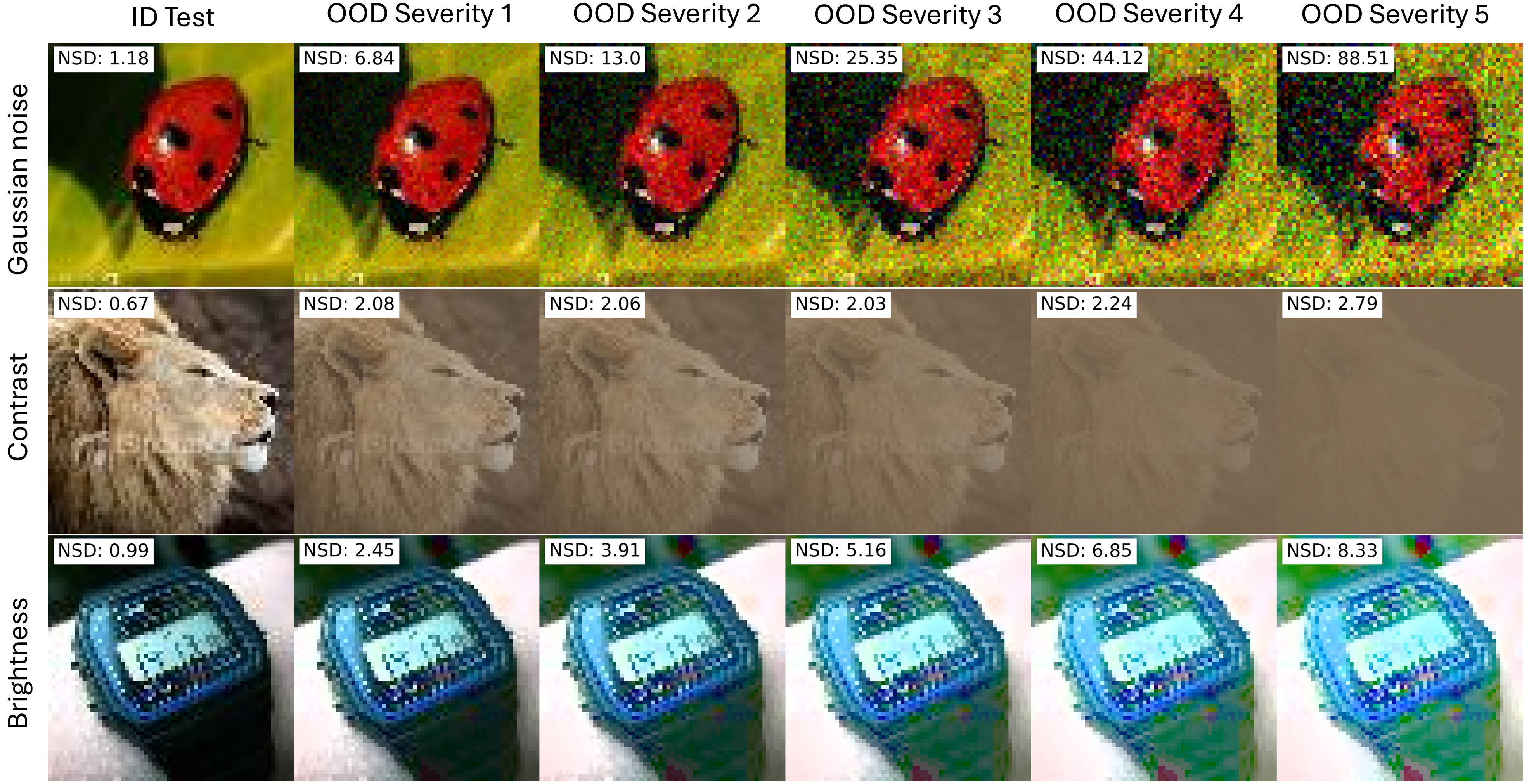}
  \caption{Example CovariateFlow~(NSD) predictions for images from the ImageNet200 ID test set and corresponding covariate shifted images from ImageNet200-C.}\label{fig:nsd_prediction_examples}
\end{figure}

\begin{wrapfigure}{r}{0.45\textwidth}
\vspace{-1cm}
  \begin{center}
    \includegraphics[width=1.0\linewidth]{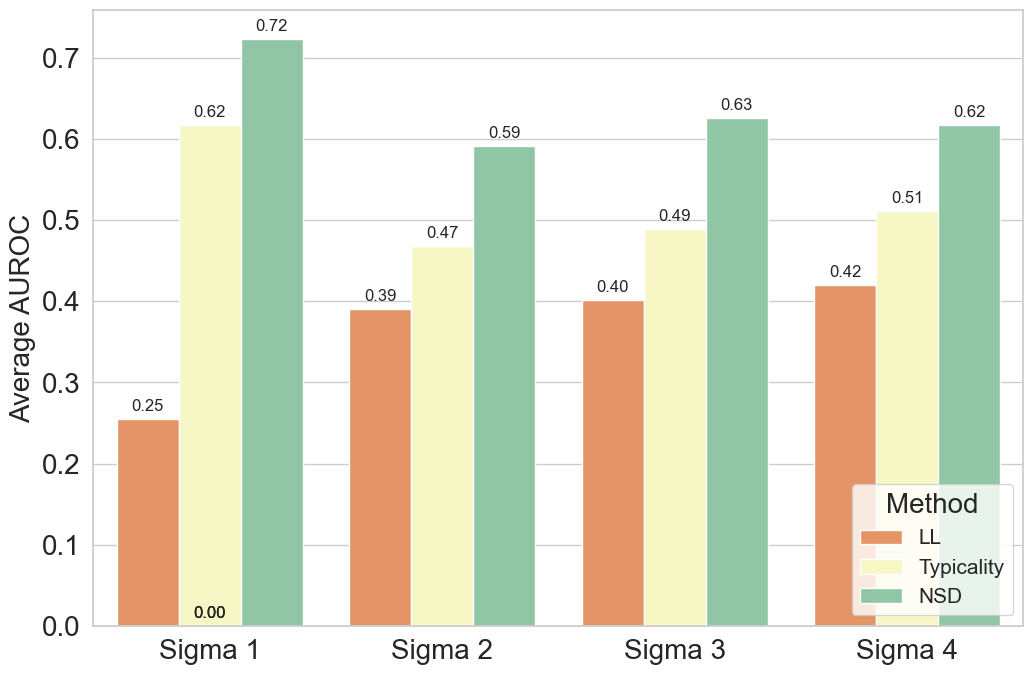}
    \caption{The average AUROC obtained with CovariateFlow on ImageNet200 vs. ImageNet200-C (all corruptions) at different filter sizes. The figure depicts the score obtained with evaluating using log-likelihood, typicality and the proposed NSD.}
    \label{fig:sigma_ablation}
  \end{center}
  \vspace{-1.0cm}
\end{wrapfigure}

To evaluate the impact of filter kernel size on performance, we conducted an experiment using CovariateFlow. Figure~\ref{fig:sigma_ablation} illustrates the average AUROC achieved with varying Gaussian filter sizes. The results indicate that a smaller filter yields the highest average performance. Example evaluations from the CovariateFlow~(NSD) model are presented in Figure~\ref{fig:nsd_prediction_examples}. Notably, the evaluated scores increase with each severity level, although the rate of increase is not linear or consistently increasing between the different degradation types. The CovariateFlow model is fully invertible and, as such, can generate heteroscedastic high-frequency components. Figure~\ref{fig:sample} depicts an example with sampled high-frequency components, the reconstructed image and a comparison between the reconstructed image and the original image. Importantly, the sampling process is stochastic and the sampling range is not limited in the example.

\newpage

\section{Discussion}\label{sec:discussion}
The findings from our analyses validate our hypothesis that OOD covariate shifts can be effectively identified by explicitly modeling the conditional distribution between low-frequency and high-frequency components. The proposed CovariateFlow, designed to specifically capture this distribution, surpasses other methodologies in detecting covariate shifts in CIFAR10 and ImageNet200. Given the diverse array of subjects and covariate conditions within the corrupted datasets, focusing on this conditional distribution streamlines the model's task, allowing it to concentrate on the most relevant distribution for the detection process.
Extending on the analysis with Table~\ref{tab:aurocresults_cifar}, the VAE-based models show adequate performance in detecting noisy degradations due to their inductive bias towards modeling low-frequency image components. On the other hand, the model falls short for this exact reason when exposed to any blurring or color degradations in the images. The DDPM with the LPIPS + MSE metric, present strong performance on noise and blurring-based covariate shift, but struggles when exposed to color shift. This is likely due to color reconstructions happening earlier in the reconstruction schedule. Consistent with existing literature~\cite{normalizingfail}, the NF-based methods evaluated using LL are extremely sensitive to noisy degradations. However, any blurring or color shift is evaluated as being highly probable under the modelled distribution, highlighting the bias of LL-based evaluation towards lower textural content. Employing the newly proposed \emph{typicality} metric shows the exact opposite behaviour. Both GLOW and the proposed CovariateFlow, fail at detecting noise-based covariate shift, but show remarkable improvements on both blurring and color-based covariate shifts when evaluated with \emph{typicality}. Combining typicality and LL in the newly proposed NSD metric, accentuates the strengths of each, enabling strong detecting performance across most of the covariates with CovariateFlow. NSD enhances the OOD detection capabilities of both the standard GLOW model and the proposed CovariateFlow, establishing it as a general and robust metric for OOD detection in NF-based models. On the higher resolution images from ImageNet200, the model also shows some effectiveness in distinguishing JPEG compression as OOD, a notoriously difficult perturbation to detect.

\textit{When to use CovariateFlow:} Despite GLOW~(LL)’s slightly superior performance in general noise detection, CovariateFlow, leveraging NSD, proves to be better overall. This provides a clear and general recommendation to the reader: LL is preferred in case strictly noise-based shifts are expected. Without a priori knowledge on the OOD shift type (which is usually the case), CovariateFlow with NSD is optimal.
This work demonstrates that it is possible to detect (even slight) perturbations in a target domain without introducing biases or prior knowledge of these perturbations into the model, unlike some contrastive learning approaches~\cite{iqa}. It only assumes access to a sufficiently large dataset that captures the \textit{in-distribution covariate shifts} and aims to detect any covariate shift outside of this distribution.
 
\section{Future Work \& Limitations}\label{sec:limitations}
Some concerns can be raised about the complexity of the \emph{typicality} computation, since test time inference requires a forward pass to compute the LL followed by a backpropagation computation per sample. This increases the memory requirements when deploying the model and decreases the overall inference speed. However, in scenarios where accurate OOD covariate shift is essential, CovariateFlow provides the best accuracy vs. speed trade-offs (see Section~\ref{sec:supp_detailed_results}). 

This work primarily focuses on detecting covariate shift, with explicit covariate shifts introduced to assess performance. Many publicly available datasets exhibit both semantic and potential covariate shifts. Although the proposed approach demonstrates effectiveness in CIFAR10 vs. SVHN (Table~\ref{tab:semantic_OOD}), future work should explore domain-specific datasets with limited ID covariate conditions to test the sensitivity of the proposed approach. As depicted in Figure~\ref{fig:nsd_prediction_examples}, the scores acquired through evaluation with CovariateFlow~(NSD) correctly increase with each severity level, however not at the same rate for each degradation type. Future work should explore the latent representations of each degradation to potentially aligning these scores with image quality metrics~\cite{koniq10k} for blind image quality assessment applications. 

\section{Conclusion}\label{sec:conclusion}
This paper explores Out-of-Distribution~(OOD) detection, specifically targeting covariate shifts caused by changes in general image statistics. This work introduces CovariateFlow, a novel approach utilizing conditional Normalizing Flows~(cNFs) for effectively targeting heteroscedastic high-frequency image components, demonstrating its superior efficacy in detecting OOD shifts across diverse datasets such as CIFAR10(-C) (74.9 \% AUROC) and ImageNet200(-C) (72.2 \% AUROC). Our analysis reveals that by meticulously modeling the conditional distribution between low-frequency and high-frequency components, CovariateFlow outperforms existing models, particularly when employing the Normalized Score Distances~(NSD) metric, which is a synthesis of log-likelihood and typicality evaluations. This approach not only highlights the importance of addressing covariate shifts for enhancing the fidelity of imaging systems, but also underscores the potential of unsupervised generative models in improving machine learning models' robustness against OOD data. 

\textbf{Acknowledgments:} This research was funded by the Philips IGTs and the Xecs Eureka TASTI Project.

\newpage

\bibliographystyle{splncs04}
\bibliography{egbib}
\newpage
\section{Supplementary }
\label{sec:supplementary}

The supplementary material is organized as follows: Section~\ref{sec:supp_implemntation_details} describes the implementation details of all the models employed in this paper. Section~\ref{sec:supp_nsd} has a step-by-step rundown on how we obtain the Normalized Score Distance. Section~\ref{sec:supp_detailed_results} provides detailed results on CIFAR10 and CIFAR10-C of the experiments and Section~\ref{sec:supp_detailed_results_tin} results on ImageNet200 and ImageNet200-C as described in the Experiments section of the main paper. Finally, we provide a series of additional ablation experiments in Section~\ref{sec:supp_ablation}.

\subsection{Implementation Details}\label{sec:supp_implemntation_details}
In this section, we detail the unsupervised training methodologies employed for five distinct baseline models and CovariateFlow aimed at OOD detection. 

\textbf{VAE and Adversersial VAE:} The VAE is trained to minimize the standard ELBO~\cite{kingma2022autoencoding} loss. Model evaluations using SSIM and KL-divergence presented the best AUROC results. The AVAE model integrates adversarial training~\cite{goodfellow2014generative} into the variational autoencoder framework to enhance its capability in generating realistic samples. For OOD detection, one can leverage the reconstruction loss (Mean Squared Error~(MSE)), the KL-divergence and the discriminative loss to compute a OOD score. We adopt the implementation described in \cite{plumerault2021avae}. In both the VAE and AVAE we employ a 4 layer deep network with a latent dim $= 1024$. The models were trained for 200 epochs following a cosine annealing learning rate scheduler.\\

\textbf{VAE-FRL:} The VAE with frequency-regularized learning~(FRL)~\cite{cai2023out} introduces decomposition and training mechanism which incorporates high-frequency information into training and guides the model to focus on semantically relevant features. This proves effective in semantic OOD detection. We employ the pretrained model as publicly available\footnote{\href{https://github.com/mu-cai/FRL/tree/main}{https://github.com/mu-cai/FRL/tree/main}}. For the CIFAR10 experiments, the model consists of a standard 3 layer deep VAE with strided convolutional down-sampling layer, transposed convolutional up-sampling and ReLu non-linear functions. The model has a latent dimension of 200. The OOD score is obtained by the log-likelihood (lower bound in the case of the VAE) minus the image complexity. The formulation is given as
\[
S(x) = -\log p_\theta(x) - L(x),
\]
where \( L(x) \) is the complexity score derived from data compressors~\cite{serra2019input}, such as PNG.\\

\textbf{Denoising Diffusion Probabilistic Model:} We implemented the Denoising Diffusion Probabilistic Model~(DDPM) following the specifications outlined in \cite{graham2023denoising} and as publicly available~\footnote{\href{https://github.com/marksgraham/ddpm-ood}{https://github.com/marksgraham/ddpm-ood}}. The method employs a time-conditioned UNet~\cite{unet} architecture with a simplified training objective where the variance is set to time-dependent constants and the model is trained to directly predict the noise $\epsilon$ at each timestep $t$:
\begin{equation}
L(\theta) = \mathbb{E}_{t,x_0,\epsilon} \left[ \|\epsilon - \epsilon_{\theta}(x_t)\|^2 \right].
\end{equation}
We aim to reconstruct an input $x_t$ across multiple time steps ($t$), utilizing the DDPM sampling strategy which necessitates $t$ steps for each reconstruction $\hat{x}_{0,}t$, with each step involving a model evaluation. To enhance efficiency, we leverage the PLMS sampler~\cite{plms}, a recent advancement in fast sampling for diffusion models, which significantly decreases the number of required sampling steps while preserving or enhancing the quality of samples. For evaluating the reconstructions, we employ both the mean-squared error~(MSE) between the reconstructed and the input image, and the Learned Perceptual Image Patch Similarity~(LPIPS) metric~\cite{zhang2018unreasonable}, the latter of which assesses perceptual similarity through deep feature distances. For each of the N reconstructions we compute these 2 similarity measurements. Finally we average these scores (over the two metrics and all the reconstructions) to derive an OOD score for each input, integrating both quantitative and perceptual accuracy assessments.

The model architecture is implemented exactly as described in \cite{graham2023denoising}. For training, we set $T=1000$ and employed a linear noise schedule, with $\beta_t$ ranging from 0.0015 to 0.0195. The training process spanned 300 epochs, utilizing the Adam optimizer with a learning rate of $2.5e^{-5}$. During the testing, we utilized the PLMS sampler configured to 100 timesteps and, in line with AnoDDPM~\cite{wyatt2022anoddpm}, we only test reconstructions from $T = 250$. Since we do not intend to detect semantic anomalies in this work and are more interested in high frequency image components, we focus on reconstructions later in the schedule. 

Finally, we experiment with the DDPM model trained on CIFAR10 and evaluated at different reconstruction starting points. Figure~\ref{fig:ddpm_results} depicts our results obtained with different reconstruction starting points and the average AUROC across all the degradations in CIFAR10-C.\\ 

\textbf{GLOW:} Normalizing Flows enable OOD detection by modeling the ID data distributions with invertible transformations through a maximize the log-likelihood training objective. We employ the GLOW~\cite{kingma2018glow} architecture, as publicly available~\footnote{\href{https://github.com/y0ast/Glow-PyTorch}{https://github.com/y0ast/Glow-PyTorch}}, in this study. Additionally, following the recent work in typicality (Section~\ref{sec:typicality}), we train our model with the \textit{approximate mass} augmented log-likelihood objective as described in~\cite{chali2023typicality}. We incorporate the \textit{approximate mass} as a component in the loss function formulation. Let $L(x; \theta) = \log(p(x; \theta))$ denote the average log-likelihood~(LL) of the model, parameterized by $\theta$, evaluated over a batch of input data $x$. Our revised training objective is expressed as:
\begin{equation}\label{eq:ll_augmented}
\min_{\theta} \left( -L(x; \theta) + \alpha \left\| \frac{\partial L(x; \theta)}{\partial x} \right\| \right)
\end{equation}
where $\alpha > 0$ signifies a hyperparameter that balances the trade-off between local enhancement of the likelihood and reduction of the gradient magnitude. We employ $\alpha = 2$ in the GLOW implementation. At test time, we compute the per sample LL and gradient score. These components are used to compute the NSD as described in Section~\ref{sec:unify}.\\

\textbf{CovariateFlow:}
Section~\ref{sec:covariate_flow} describes the CovariateFlow model proposed in this work. Figure~\ref{fig:covariateflow_arch} depicts the architecture and general flow of information during training and when computing the OOD scores. Figure~\ref{fig:coupling} illustrates a detailed diagram of the low-frequency conditioned coupling steps employed in the model. Additionally, following the image decomposition through the Gaussian filtering, we encode the individual components as 16-bit depth data to avoid information loss. Our model is completely invertible and can thus also generate signal-dependent high-frequency components. The models are prepared following the typicality augmented training objective (Equation~\ref{eq:ll_augmented}). We use an Adam optimizer (starting $lr = 5e^{-4}$) with a one-cycle annealing learning rate scheduler for 300 epochs across all our experiments. The code for the model is available at \href{https://github.com/cviviers/CovariateFlow}{https://github.com/cviviers/CovariateFlow}.

\subsection{Detailed analysis of the normalized score distance (NSD)}\label{sec:supp_nsd}

\begin{figure}[!htbp]
\centering
\begin{subfigure}{.7\textwidth}
 \centering
 \includegraphics[width=1.0\linewidth]{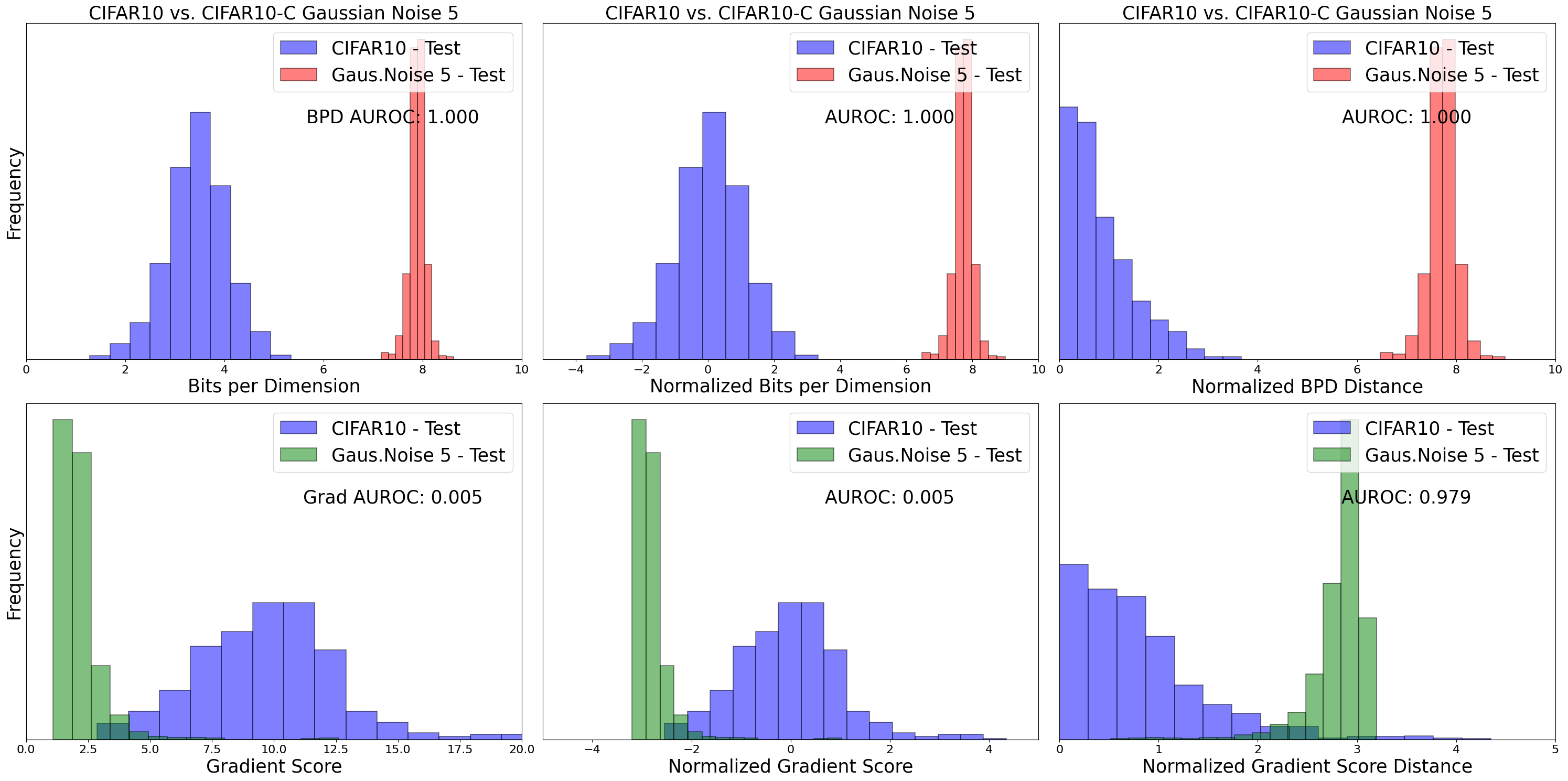}
 \caption{Top row: \textbf{Log-likelihood} of CIFAR10 vs. Gaussian Noise 5, the normalized LL and the absolute value of the normalized LL.
 \newline Bottom row: \textbf{Gradient score} of CIFAR10 vs. CIFAR10-C Gaussian Noise 5, normalized gradient score and the absolute value of the normalized gradient score.}
 \label{fig:noise_ll_tp}
\end{subfigure}%
\hspace{0.2cm}
\begin{subfigure}{.25\textwidth}
 \centering
 \includegraphics[width=0.9\linewidth]{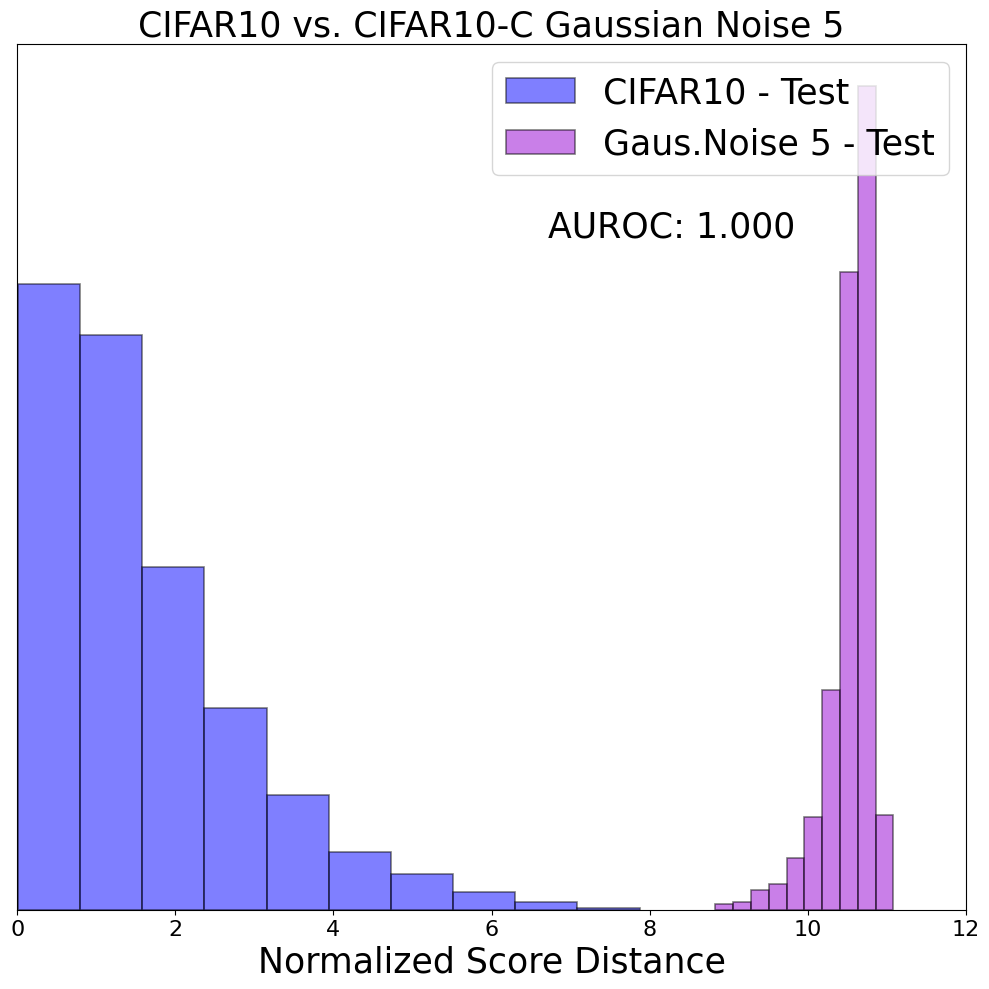}
 \caption{The sum of the normalized LL distance and the normalized gradient distance shown as a unified normalized score distance~(NSD)}
 \label{fig:noise_nsd}
\end{subfigure}

\vspace{0.5cm}
\begin{subfigure}{.7\textwidth}
 \centering
 \includegraphics[width=0.95\linewidth]{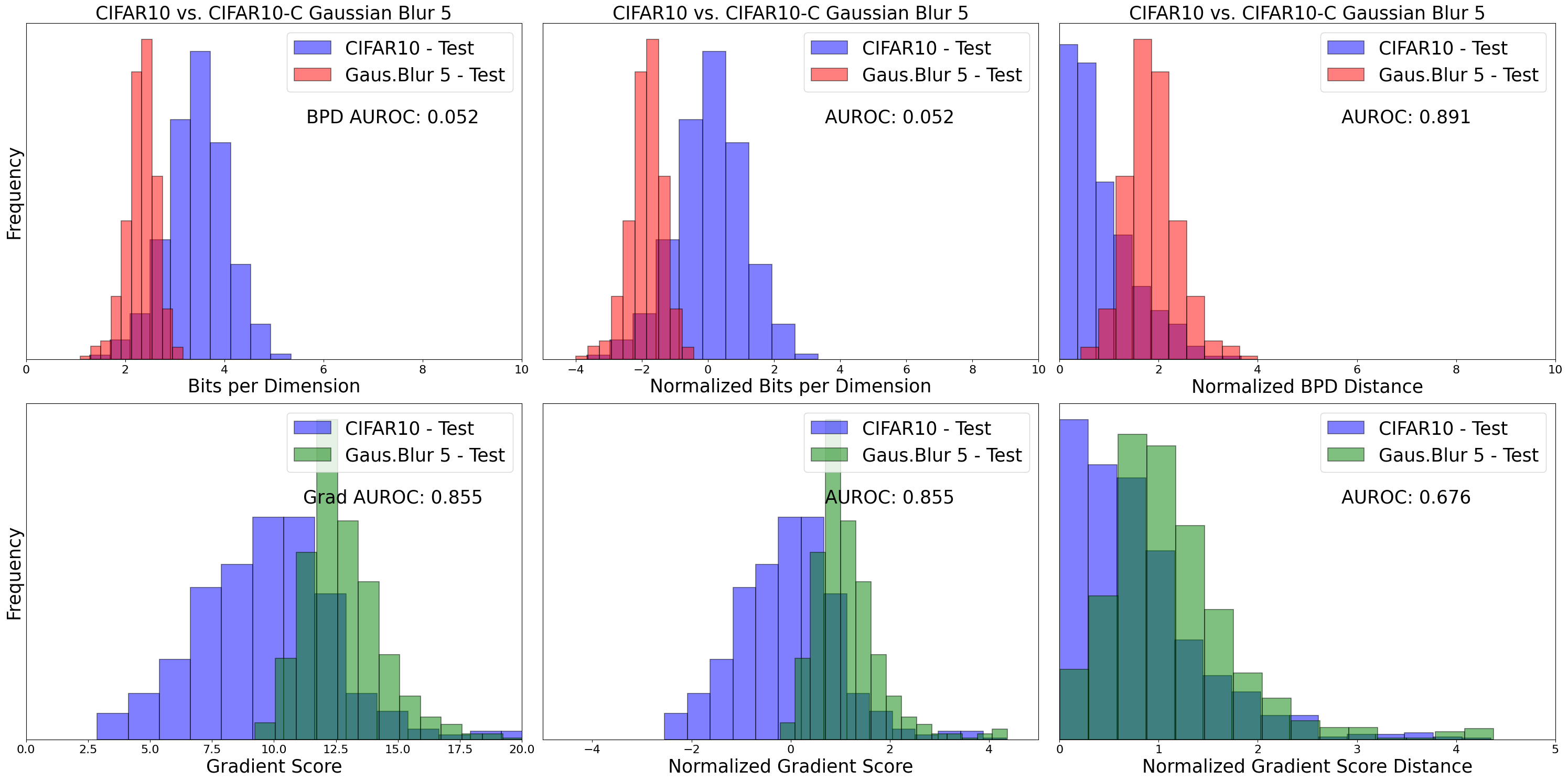}
 \caption{Top row: \textbf{Log-likelihood} of CIFAR10 vs. Gaussian Blur 5, the normalized LL and the absolute value of the normalized LL. \newline Bottom row: \textbf{Gradient score} of CIFAR10 vs. Gaussian Blur 5, normalized gradient score and the absolute value of the normalized gradient score.}
 \label{fig:blur_ll_tp}
\end{subfigure}%
\hspace{0.2cm}
\begin{subfigure}{.25\textwidth}
 \centering
 \includegraphics[width=0.95\linewidth]{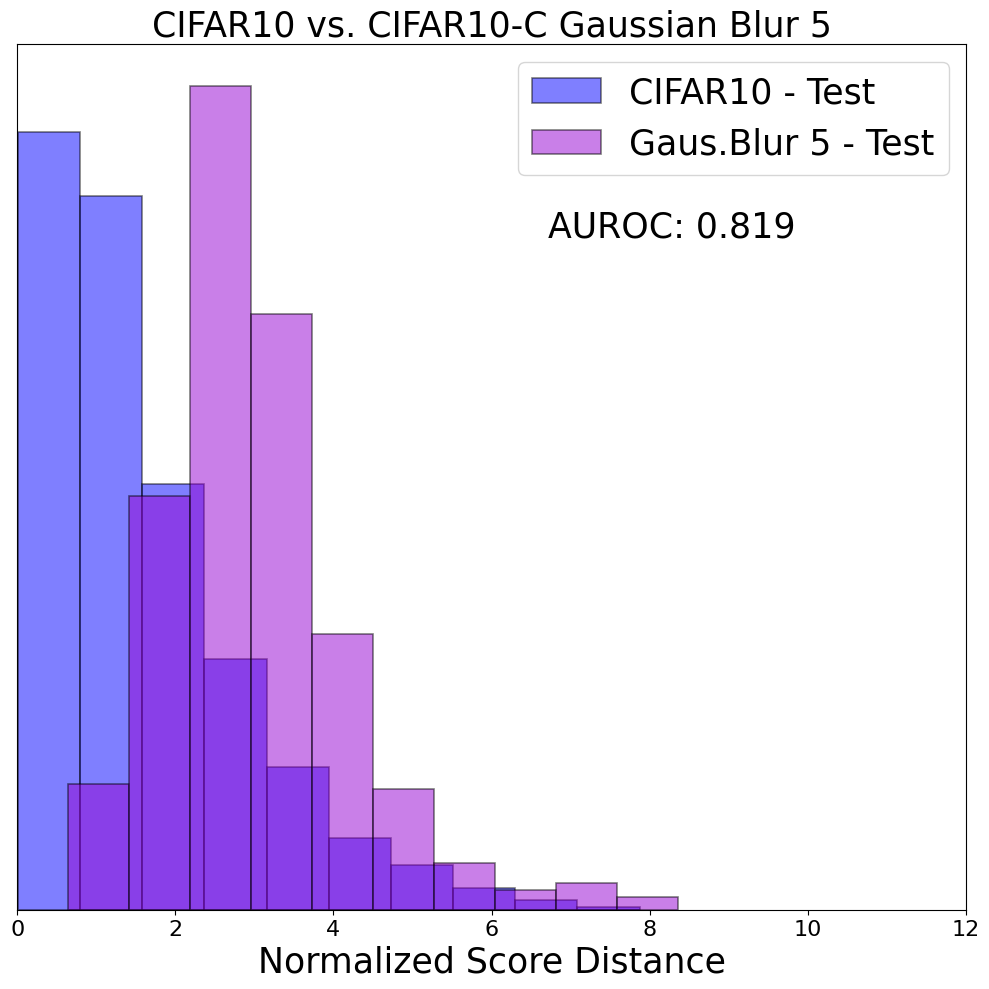}
 \caption{The sum of the normalized LL distance and the normalized gradient distance shown as a unified normalized score distance~(NSD)}
 \label{fig:blur_nsd}
\end{subfigure}

\vspace{0.5cm}
\begin{subfigure}{.7\textwidth}
 \centering
 \includegraphics[width=0.95\linewidth]{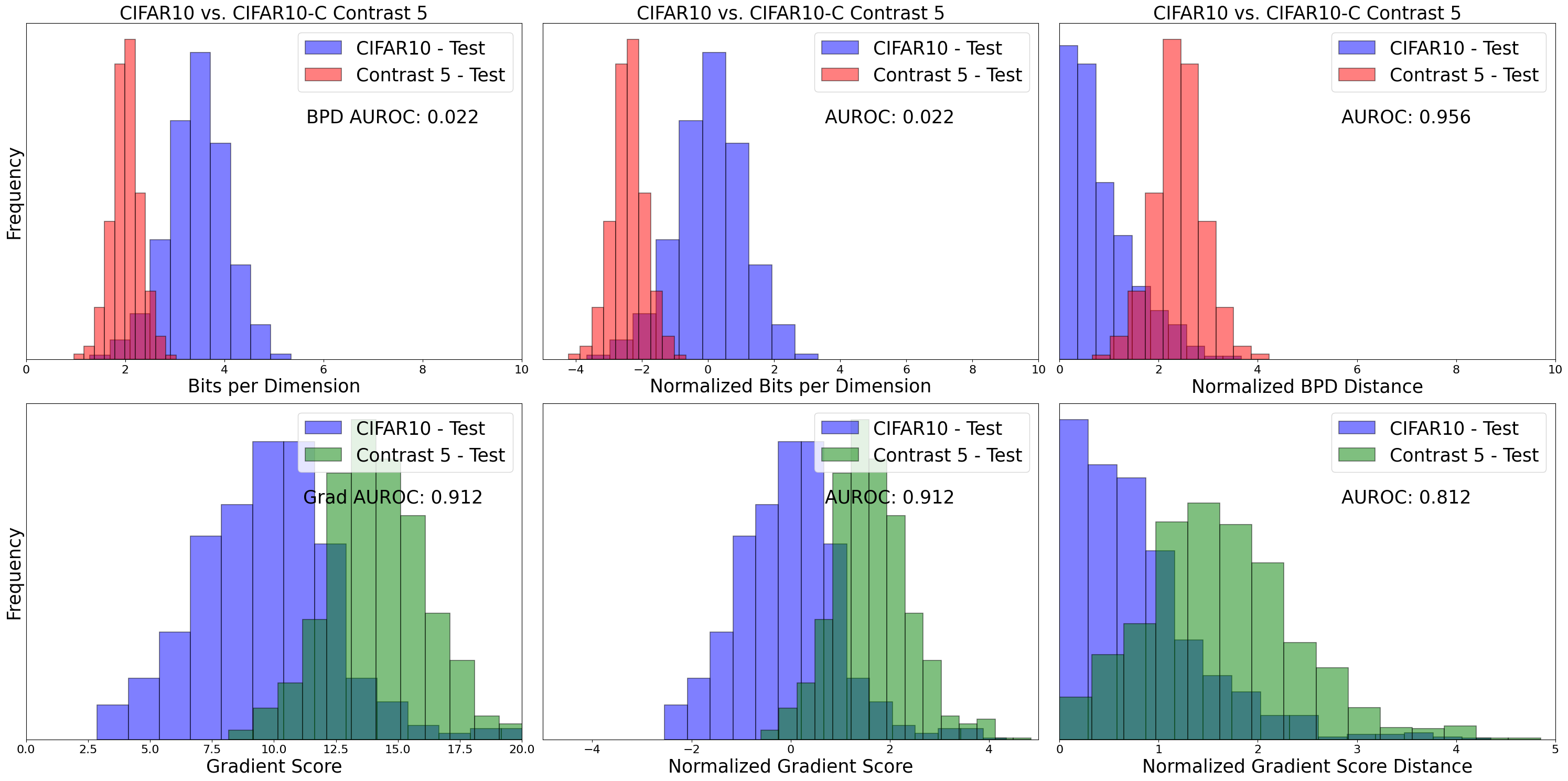}
 \caption{Top row: \textbf{Log-likelihood} of CIFAR10 vs. Contrast 5, the normalized LL and the absolute value of the normalized LL. \newline Bottom row: \textbf{Gradient score} of CIFAR10 vs. Contrast 5, normalized gradient score and the absolute value of the normalized gradient score.}
 \label{fig:Contrast_ll_tp}
\end{subfigure}%
\hspace{0.2cm}
\begin{subfigure}{.25\textwidth}
 \centering
 \includegraphics[width=0.95\linewidth]{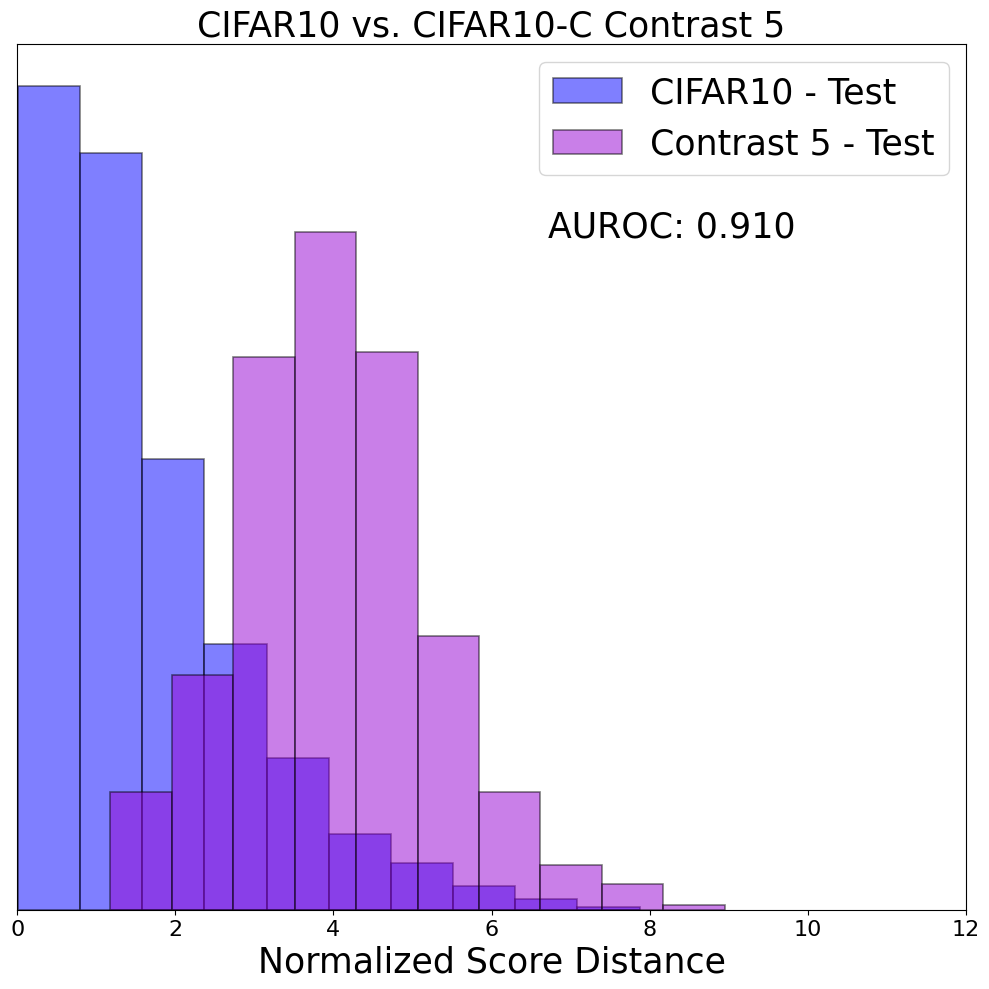}
 \caption{The sum of the normalized LL distance and the normalized gradient distance shown as a unified normalized score distance~(NSD)}
 \label{fig:Contrast_nsd}
\end{subfigure}

\caption{Histograms of test results of GLOW trained on CIFAR10 and evaluated on CIFAR10-C Gaussian Noise, Gaussian Blur and Contrast. The unification between log-likelihood and typicality to compute the Normalized Score Distance~(NSD) is depicted.}

\label{fig:nsd}
\end{figure}

This section details the computation of the NSD from the LL and typicality score. Figure~\ref{fig:nsd} depicts this process through the evaluation of the GLOW model applied to three different OOD covariate shifts. In Figure~\ref{fig:noise_ll_tp} the LL and typicality (gradient score) of the model subject to Gaussian Noise can be seen. Following the process described in Section~\ref{sec:unify}, column 2 depicts the standardization of both scores using validation statistics. This is followed by converting the scores to absolute distance from the expected mean in column 3. The LL distance and gradient score distance can then simply be added to obtain a unified distance (Figure~\ref{fig:noise_ll_tp}).
The same flow is depicted in Figure~\ref{fig:blur_ll_tp} and Figure~\ref{fig:blur_nsd} for the model subject to Gaussian Blur and Figure~\ref{fig:Contrast_ll_tp} and Figure~\ref{fig:Contrast_nsd} for Contrast change. Following this standardized approach, the change in each measure (LL and gradient score) w.r.t. the validation statistics are utilized and combined to provide a single and effective OOD score. All the results depicted in The Figure~\ref{fig:nsd} depicts the ID CIFAR10 test scores vs. the OOD CIFAR10-C scores.

\newpage
\subsection{Detailed Results on CIFAR10 vs. CIFAR10-C}\label{sec:supp_detailed_results}
The following section presents detailed results obtained with various models on our experiments with ID CIFAR10 and CIFAR10-C as OOD.

\begin{figure}[!htbp]
\begin{minipage}[b]{1.0\linewidth}
 \centering
 \centerline{\includegraphics[width=6.5cm]{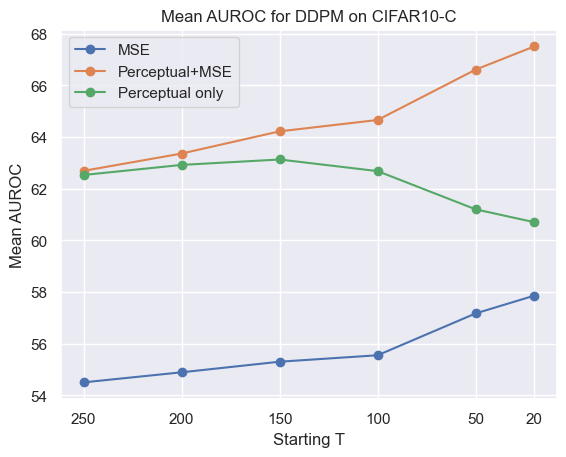}}
\end{minipage}
\caption{Results obtained with the DDPM on CIFAR10 and CIFAR10-C. The figure depicts mean AUROC obtained from reconstructions at different starting points, T. With covariate shift being predominantly change in high-frequency components, reconstructions starting at T=20 shows the best performance. }
\label{fig:ddpm_results}
\vspace{-0.5cm}
\end{figure}

Our analysis examines the reconstruction capabilities of the DDPM across various initial time steps, $T$. Figure~\ref{fig:ddpm_results} presents the mean AUROC curve calculated for reconstructions assessed using the LPIPS, MSE, or a combination of LPIPS and MSE metrics at each time step. Notably, at larger time steps (e.g., $T=250$), the distinction in average reconstruction error between the ID CIFAR10 test set and the OOD CIFAR10-C dataset becomes less pronounced, leading to inferior OOD detection performance. This phenomenon is attributable to the high-level image perturbations characteristic of OOD data, which are predominantly addressed in the final stages of the diffusion process. In Contrast, initial diffusion stages focus on generating lower-level image semantics, resulting in reconstructions that significantly diverge from the test image, particularly in terms of low-frequency components.

Figures~\ref{fig:nsd}, \ref{fig:ddpm_results}, and Table~\ref{tab:aurocresults_cifar}, highlight the distinct sensitivities of log-likelihood (LL) and gradient scores when applied to GLOW under severe Gaussian Noise conditions, as depicted in Figure~\ref{fig:noise_ll_tp}. These metrics diverge in their assessment, with LL clearly identifying distorted images as OOD, whereas gradient scores suggest such images are more typical than even the ID data. Conversely, Figure~\ref{fig:blur_ll_tp} demonstrates the opposite trend for blurred images, where LL overestimates their likelihood relative to ID data, but gradient scores accurately classify them as OOD. These observations corroborate Zhang~\etal's theoretical insights~\cite{zhang2021understandingfailuresOOD} about the propensity of certain model-metric combinations to misjudge the probability of natural images. To address these discrepancies, we introduce the NSD metric, which synthesizes LL and gradient movements into a unified OOD detection metric. Figures~\ref{fig:noise_nsd}, \ref{fig:blur_nsd} and \ref{fig:Contrast_nsd} validate the NSD metric's effectiveness in discerning OOD samples across both conditions, with extended results available in the supplementary material.
\begin{table*}[htbp]
\centering
{\resizebox{1.0\linewidth}{!}{
\begin{tabular}{lcccc} 
\toprule
 \multicolumn{1}{c}{\underline{ CIFAR10 ID } }      &              \multicolumn{4}{c}{\underline{CIFAR10-C OOD }}\\
 \enspace &\enspace Gaussian Noise \enspace &\enspace Gaussian Blur \enspace &\enspace Contrast \enspace & \enspace All Shifts \\
Method & 1/ 2/3/4/5 & 1/ 2/3/4/5 & 1/ 2/3/4/5 & Average$\uparrow$ FPR95$\downarrow$\\
\midrule
\textbf{Reconstruction} \\
\midrule
DDPM~\cite{graham2023denoising} (T150: LPIPS)& 53.7/ 59.2/ 66.3/ 70.1/ 73.5 \enspace & \enspace 50.7/ 68.3/ 82.4/ 92.1/ 98.8 \enspace & \enspace 50.1/ 50.0/ 51.2/ 50.5/ 50.2 & 63.1/ 83.9\\ 
DDPM~\cite{graham2023denoising} (T20: LPIPS + MSE) & 75.6/ 91.7/ 98.2/ 99.1/ 99.6\enspace & \enspace 48.7/ 58.6/ 70.6/ 82.2/ 95.1 \enspace & \enspace 48.2/ 48.5/ 48.3/ 46.2/ 45.0 &  67.5/ 75.2 \\
\midrule
\textbf{Explicit Density} \\
\midrule
Vanilla VAE~\cite{kingma2022autoencoding} (SSIM + KL Div) & 64.2/ 79.0/ 91.7/ 95.6/ 97.9 & \enspace 43.0/ 24.6/ 19.2/ 15.4/ 10.2 & 23.8/ 5.4/ 2.0/ 0.5/ 0.0 &  48.9/ 83.3\\
AVAE~\cite{plumerault2021avae} (MSE + KL Div + Adv Loss) & 58.4/ 68.7/ 80.6/ 86.1/ 90.6 & 45.5/ 34.0/ 30.7/ 28.2/ 25.7 & 34.1/ 38.3/ 43.5/ 48.3/ 50.3 & 60.2/ 73.1\\
GLOW~\cite{kingma2018glow} (LL) & 100.0/ 100.0/ 100.0/ 100.0/ 100.0 & 44.3/ 21.3/ 14.2/ 9.8/ 5.2 & 39.2/ 20.9/ 14.9/ 8.8/ 2.2/ & 57.7/69.5\\
GLOW~\cite{kingma2018glow} (Typicality)~\cite{chali2023typicality} & 0.0/ 0.0/ 0.2/ 0.2/ 0.47 & 55.4/ 65.1/ 71.1/ 76.5/ 85.53 & 60.4/ 66.1/ 71.5/ 77.7/ 91.2 &  41.6/85.8\\
GLOW (Normalized Distance) & 100.0/ 100.0/ 100.0/ 100.0/ 100.0 & 48.7/ 52.7/ 60.8/ 69.2 / 82.0 & 49.6/ 57.7/ 64.4/ 74.0/ 91.0 &  69.3/65.7\\
CovariateFlow (LL) & 100.0/ 100.0/ 100.0/ 100.0/ 100.0 \enspace & \enspace 42.1/ 16.4/ 10.5/ 7.4/ 4.4 \enspace & \enspace 31.2/ 11.0/ 6.3/ 2.8/ 0.5 & 58.3/63.5\\
CovariateFlow (Typicality)~\cite{chali2023typicality} & 7.0/ 1.8/ 0.4/ 0.2/ 0.1\enspace & \enspace 56.1/ 75.9/ 81.0/ 84.6/ 89.5\enspace & \enspace63.3/ 77.6/ 81.9/ 85.8/ 91.1 \enspace &  45.5/83.8\\
CovariateFlow (NSD) & 99.5/ 99.7/ 99.8/ 99.8/ 99.8 \enspace & 50.1/ 69.4/ 77.3/ 82.7/ 89.4 \enspace & \enspace 55.3/ 76.7/ 83.9/ 90.1/ 95.9 \enspace &  \textbf{74.9}/\textbf{61.7}\\
\bottomrule 
\end{tabular}}
\caption{AUROC scores of various methods on detecting OOD covariate shift on CIFAR10 vs. CIFAR10-C. Note, only 3 degradations at the 5 severity levels are depicted but the average AUROC and FPR95 is computed across all degradations in the dataset.\label{tab:aurocresults_cifar} }}
\vspace{-0.5cm}
\end{table*}

\begin{figure}[!htbp]

\begin{minipage}[b]{1.0\linewidth}
 \centering
 \centerline{\includegraphics[width=9.5cm]{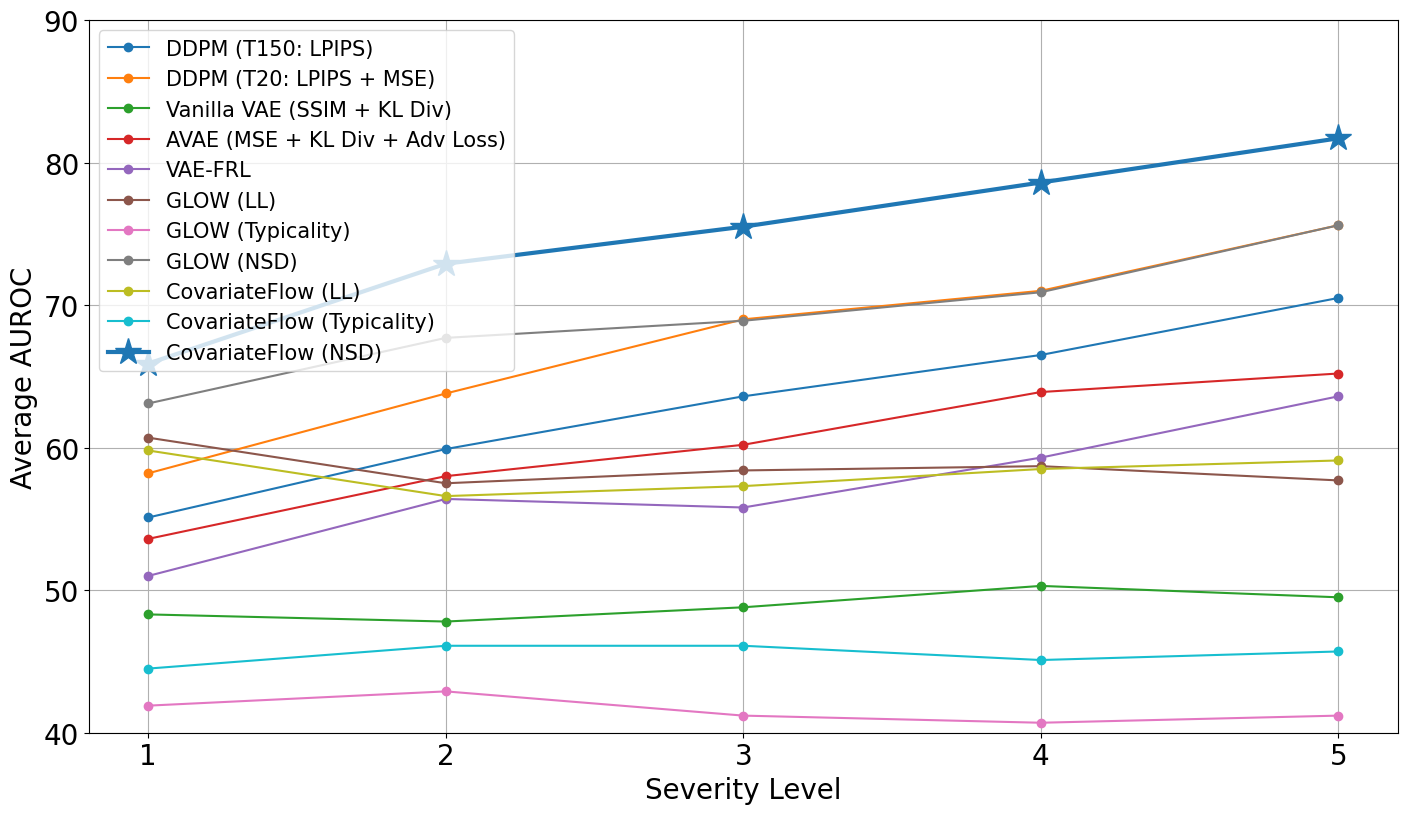}}
\end{minipage}
\caption{Illustration of model performance (average AUROC across all 19 degradations) per severity level.}
\label{fig:auroc_per_level}
\end{figure}

Table~\ref{tab:aurocresults_cifar} depicts the AUROC for 3 degradations (each severity) from CIFAR-10C that summarizes the performance of all the models employed in this work. Figure~\ref{fig:auroc_per_level} additionally depicts the average AUROC of all the models at each severity. We also present the complete performance evaluation of all the models on CIFAR10-C on all the degradtions and at every severity level. The results are depicted in order of presentation: DDPM T150-LPIPS (\ref{tab:ddpm_results_LPIPS}), DDPM T20-LPIPS+MSE~(\ref{tab:ddpm_results}), VAE~(Table~\ref{tab:vae_results}), AVAE~(Table~\ref{tab:avae_results}), GLOW-LL~(Table~\ref{tab:glow_results_ll}), GLOW-Typicality~(Table~\ref{tab:glow_results_typ}), GLOW-NSD~(Table~\ref{tab:glow_results_nsh}), CovariateFlow-LL~(Table~\ref{tab:covariate_results_ll}), CovariateFlow-Typicality~(Table~\ref{tab:covariate_results_typ}) and CovariateFlow-NSD~(Table~\ref{tab:covariate_results_nsd}).

\begin{table*}[htbp]
\centering
{\resizebox{1.0\linewidth}{!}{
\begin{tabular}{r|cccccc}
Severity & 1 & 2 & 3 & 4 & 5 & Average \\
Metric & AUROC$\uparrow$/ FPR95$\downarrow$ & AUROC$\uparrow$/ FPR95$\downarrow$ & AUROC$\uparrow$/ FPR95$\downarrow$ & AUROC$\uparrow$/ FPR95$\downarrow$ & AUROC$\uparrow$/ FPR95$\downarrow$ & AUROC$\uparrow$/ FPR95$\downarrow$ \\
\midrule
Gaussian Noise & 53.74 / 95.2 & 59.17 / 93.3 & 66.26 / 89.5 & 70.09 / 88.6 & 73.54 / 84.0 & 64.56 / 90.12 \\
Shot Noise & 52.78 / 95.8 & 53.83 / 95.2 & 61.61 / 91.9 & 65.4 / 88.0 & 72.95 / 81.8 & 61.31 / 90.54 \\
Speckle Noise & 52.19 / 96.0 & 52.19 / 96.0 & 59.51 / 92.6 & 67.72 / 86.5 & 74.94 / 77.8 & 61.31 / 89.78\\
Impulse Noise & 61.15 / 92.2 & 68.9 / 86.1 & 76.25 / 77.1 & 86.74 / 53.6 & 91.72 / 39.6 & 76.95 / 69.72 \\
Defocus Blur & 50.17 / 96.3 & 54.95 / 93.7 & 67.72 / 89.4 & 81.99 / 70.6 & 96.95 / 14.1 & 70.36 / 72.82 \\
Gaussian Blur & 50.71 / 96.3 & 68.3 / 86.5 & 82.39 / 68.6 & 92.07 / 39.8 & 98.81 / 5.0 & 78.46 / 59.24\\
Glass Blur& 64.67 / 85.1 & 63.36 / 85.1 & 57.20 / 87.1 & 73.2 / 89.3 & 66.29 / 77.6 & 64.94 / 84.84\\
Motion Blur & 60.81 / 92.1 & 74.48 / 77.3 & 83.69 / 63.2 & 84.0 / 66.7 & 90.22 / 45.9 & 78.64 / 69.04 \\
Zoom Blur & 71.56 / 80.9 & 74.32 / 76.4 & 80.52 / 70.9 & 84.12 / 61.8 & 89.68 / 49.0 & 80.04 / 67.8 \\
Snow & 51.64 / 96.3 & 52.8 / 95.8 & 51.38 / 95.2 & 48.4 / 96.3 & 46.46 / 95.3 & 50.14 / 95.78 \\
Fog & 56.53 / 93.9 & 70.66 / 80.8 & 81.27 / 61.4 & 89.63 / 39.6 & 94.99 / 20.7 & 78.62 / 59.28\\
Brightness & 50.2 / 96.0 & 48.75 / 96.4 & 47.56 / 96.9 & 46.16 / 96.9 & 44.4 / 97.4 & 47.41 / 96.72 \\
Contrast & 50.12 / 95.2 & 49.97 / 94.7 & 51.18 / 95.0 & 50.49 / 92.8 & 50.19 / 95.0 & 50.39 / 94.54\\
Elastic Transform & 57.55 / 93.3 & 58.12 / 92.7 & 63.66 / 88.6 & 60.5 / 89.1 & 53.99 / 93.5 & 58.76 / 91.44 \\
Pixelate & 52.0 / 94.4 & 53.44 / 93.2 & 54.06 / 93.3 & 58.04 / 89.5 & 64.96 / 85.9 & 56.5 / 91.26 \\
JPEG Compression & 54.9 / 93.9 & 56.57 / 93.0 & 57.6 / 92.7 & 57.65 / 93.0 & 60.04 / 90.2 & 57.35 / 92.56 \\
Spatter & 55.48 / 93.5 & 60.96 / 92.6 & 61.25 / 89.4 & 56.48 / 90.3 & 63.88 / 86.1 & 59.61 / 90.38 \\
Saturate & 64.44 / 90.7 & 73.69 / 84.1 & 45.51 / 97.1 & 40.4 / 98.0 & 36.85 / 98.2 & 52.18 / 93.62 \\
Frost & 46.57 / 97.0 & 47.15 / 97.0 & 54.29 / 94.6 & 56.36 / 95.0 & 64.77 / 91.6 & 53.83 / 95.04 \\
\bottomrule
\end{tabular}}

\caption{The performance of the Denoising Diffusion Probabilistic Model (DDPM) in detecting out-of-distribution (OOD) covariate shift between CIFAR10 and CIFAR10-C datasets is evaluated. The model is evaluated with a starting \textbf{T=150 and using the LPIPS reconstruction} metric. The model achieves a mean Area Under the Receiver Operating Characteristic (AUROC) of 63.2\% and a False Positive Rate at 95\% True Positive Rate (FPR95) of 84.0\%.\label{tab:ddpm_results_LPIPS}}}
\end{table*}

\begin{table*}[htbp]
\centering
{\resizebox{1.0\linewidth}{!}{
\begin{tabular}{r|cccccc}
Severity & 1 & 2 & 3 & 4 & 5 & Average\\
Metric & AUROC$\uparrow$/ FPR95$\downarrow$ & AUROC$\uparrow$/ FPR95$\downarrow$ & AUROC$\uparrow$/ FPR95$\downarrow$ & AUROC$\uparrow$/ FPR95$\downarrow$ & AUROC$\uparrow$/ FPR95$\downarrow$ & AUROC$\uparrow$/ FPR95$\downarrow$  \\
\midrule
Gaussian Noise & 75.55 / 79.1 & 91.74 / 30.8 & 98.2 / 5.0 & 99.14 / 2.2 & 99.57 / 1.3 & 92.84 / 23.68 \\
Shot Noise & 67.34 / 85.8 & 79.52 / 65.6 & 95.65 / 17.2 & 97.69 / 8.0 & 99.23 / 2.1 & 87.89 / 35.74 \\
Speckle Noise & 68.09 / 79.8 & 68.09 / 79.8 & 93.4 / 25.6 & 97.77 / 8.2 & 99.2 / 2.3 & 85.31 / 39.14 \\
Impulse Noise & 88.62 / 41.7 & 98.31 / 5.0 & 99.64 / 1.0 & 99.99 / 0.1 & 100.0 / 0.0 & 97.31 / 9.56 \\
Defocus Blur & 48.79 / 94.6 & 49.96 / 95.5 & 57.66 / 92.9 & 70.48 / 88.9 & 91.46 / 49.2 & 63.67 / 84.22 \\
Gaussian Blur & 48.65 / 95.4 & 58.56 / 93.4 & 70.64 / 87.6 & 82.24 / 76.6 & 95.11 / 30.3 & 71.04 / 76.66 \\
Glass Blur & 75.7 / 79.2 & 73.6 / 80.6 & 64.93 / 86.5 & 79.38 / 76.6 & 71.9 / 81.0 & 73.1 / 80.78 \\
Motion Blur & 54.68 / 93.4 & 64.04 / 91.5 & 72.44 / 86.0 & 72.64 / 88.1 & 79.69 / 81.8 & 68.7 / 88.16 \\
Zoom Blur & 62.21 / 91.4 & 65.0 / 89.7 & 69.71 / 85.3 & 73.18 / 84.2 & 78.89 / 79.5 & 69.8 / 86.02 \\
Snow & 58.47 / 93.8 & 66.97 / 89.0 & 63.95 / 91.3 & 59.71 / 94.4 & 56.71 / 96.7 & 61.16 / 93.04 \\
Fog & 54.46 / 92.1 & 62.74 / 86.0 & 69.58 / 77.9 & 77.22 / 73.8 & 86.11 / 54.0 & 70.02 / 76.76 \\
Brightness & 52.34 / 94.9 & 51.03 / 95.5 & 51.96 / 95.9 & 51.58 / 96.2 & 50.44 / 97.5 & 51.47 / 96.0 \\
Contrast & 48.24 / 95.4 & 48.47 / 95.4 & 48.31 / 95.6 & 46.2 / 95.9 & 44.95 / 95.9 & 47.23 / 95.64\\
Elastic Transform & 52.07 / 94.0 & 52.03 / 93.8 & 55.79 / 93.8 & 52.77 / 93.1 & 50.06 / 93.7 & 52.54 / 93.68 \\
Pixelate & 50.41 / 95.4 & 54.13 / 93.9 & 53.29 / 93.4 & 57.41 / 91.5 & 61.82 / 91.1 & 55.41 / 93.06 \\
JPEG Compression & 54.49 / 93.8 & 55.73 / 92.4 & 56.5 / 92.8 & 56.4 / 91.5 & 58.97 / 90.4 & 56.42 / 92.18 \\
Spatter & 57.57 / 91.6 & 66.94 / 83.2 & 72.93 / 79.8 & 63.86 / 84.4 & 76.67 / 67.9 & 67.59 / 81.38 \\
Saturate & 53.83 / 95.9 & 61.16 / 94.6 & 51.18 / 95.4 & 55.59 / 92.2 & 61.01 / 91.5 & 56.55 / 93.92 \\
Frost & 51.76 / 96.5 & 54.24 / 97.3 & 61.21 / 95.4 & 63.83 / 94.9 & 70.32 / 86.5 & 60.27 / 94.12 \\
\bottomrule
\end{tabular}}
\caption{The performance of the Denoising Diffusion Probabilistic Model (DDPM) in detecting out-of-distribution (OOD) covariate shift between CIFAR10 and CIFAR10-C datasets is evaluated. The model is evaluated with a starting \textbf{T=20 and using the MSE + LPIPS reconstruction} metric. The model achieves a mean Area Under the Receiver Operating Characteristic (AUROC) of 67.8\% and a False Positive Rate at 95\% True Positive Rate (FPR95) of 75.5\%.\label{tab:ddpm_results}}}
\end{table*}

\begin{table*}[htbp]
\centering
{\resizebox{1.0\linewidth}{!}{
\begin{tabular}{r|cccccc}
Severity & 1 & 2 & 3 & 4 & 5 & Average \\
Metric & AUROC$\uparrow$/ FPR95$\downarrow$ & AUROC$\uparrow$/ FPR95$\downarrow$ & AUROC$\uparrow$/ FPR95$\downarrow$ & AUROC$\uparrow$/ FPR95$\downarrow$ & AUROC$\uparrow$/ FPR95$\downarrow$ & AUROC$\uparrow$/ FPR95$\downarrow$ \\
\midrule
Gaussian Noise & 64.2 / 82.2 & 79.0 / 58.6 & 91.7 / 26.9 & 95.6 / 15.0 & 97.9 / 7.4 & 85.68 / 38.02 \\
Shot Noise & 58.4 / 88.7 & 66.4 / 79.3 & 84.5 / 48.3 & 90.4 / 33.9 & 96.3 / 14.7 & 79.2 / 52.98 \\
Speckle Noise & 58.8 / 88.8 & 72.4 / 74.9 & 79.4 / 64.6 & 90.1 / 42.7 & 95.6 / 22.8 & 79.26 / 58.76 \\
Impulse Noise & 75.9 / 67.6 & 91.0 / 33.0 & 97.2 / 11.6& 99.8 / 0.9& 100.0 / 0.1 & 92.78 / 22.64 \\
Defocus Blur & 42.8 / 96.2& 32.2 / 97.6& 24.6 / 98.5& 20.2 / 98.8& 13.1 / 99.4 & 26.58 / 98.1 \\
Gaussian Blur & 43.0 / 96.2& 24.6 / 98.5& 19.2 / 98.9& 15.4 / 99.3& 10.2 / 99.6 & 22.48 / 98.5 \\ 
Glass Blur & 60.9 / 92.4 & 58.50 / 93.9 & 44.30 / 96.2 & 64.34 / 92.2 & 49.42 / 95.8 & 55.49 / 94.1 \\
Motion Blur & 31.8 / 97.7& 24.0 / 98.6& 18.6 / 99.1& 18.6 / 99.1& 14.8 / 99.4 & 21.56 / 98.78 \\
Zoom Blur & 27.5 / 98.2 & 23.9 / 98.4 & 21.0 / 98.6 & 18.5 / 98.8 & 15.7 / 99.1 & 21.32 / 98.62 \\ 
Snow & 58.2 / 89.4& 65.8 / 81.2& 71.1 / 75.6& 70.2 / 75.6& 65.3 / 83.7 & 66.12 / 81.1 \\
Fog & 31.9 / 98.1& 16.6 / 99.4& 10.2 / 99.6& 7.0 / 99.7& 4.2 / 99.7 & 13.98 / 99.3 \\
Brightness & 52.4 / 94.6& 54.6 / 93.5& 56.7 / 92.6& 58.0 / 92.4& 59.0 / 92.4 & 56.14 / 93.1 \\
Contrast & 23.8 / 98.8& 5.4 / 99.8& 2.0 / 99.9& 0.5 / 100.0& 0.0 / 100.0 & 6.34 / 99.7 \\
Elastic Transform & 37.3 / 97.2& 33.5 / 97.7& 27.6 / 98.2& 27.2 / 98.4& 30.4 / 98.0 & 31.2 / 97.9 \\ 
Pixelate & 49.0 / 95.3& 48.1 / 95.6& 47.5 / 95.8& 45.9 / 96.0& 43.1 / 96.3 & 46.72 / 95.8 \\
JPEG Compression & 49.7 / 95.1& 48.5 / 95.5& 47.9 / 95.6& 47.4 / 95.8& 46.7 / 95.7 & 48.04 / 95.54 \\
Spatter & 57.6 / 89.7& 68.4 / 76.8& 77.3 / 61.6& 73.8 / 78.1& 83.2 / 66.1 & 72.06 / 74.46 \\
Saturate & 42.4 / 96.7& 42.4 / 96.9& 59.1 / 91.9& 69.1 / 86.3& 76.3 / 82.1 & 57.86 / 90.78 \\
Frost & 51.8 / 93.6& 53.1 / 92.7& 48.1 / 93.1& 44.5 / 94.2& 38.4 / 95.2 & 47.18 / 93.76 \\
\bottomrule
\end{tabular}}

\caption{The performance of the VAE model in detecting out-of-distribution (OOD) covariate shift between CIFAR10 and CIFAR10-C datasets is evaluated. The model is evaluated using \textbf{MSE + KL-divergence + Adversarial} as metric. The model achieves a mean Area Under the Receiver Operating Characteristic (AUROC) of 48.9\% and a False Positive Rate at 95\% True Positive Rate (FPR95) of 83.3\%.}\label{tab:vae_results}}
\end{table*}

\begin{table*}[htbp]
\centering
{\resizebox{1.0\linewidth}{!}{
\begin{tabular}{r|ccccccc}
Severity & 1 & 2 & 3 & 4 & 5 & Average \\
Metric & AUROC$\uparrow$/ FPR95$\downarrow$ & AUROC$\uparrow$/ FPR95$\downarrow$ & AUROC$\uparrow$/ FPR95$\downarrow$ & AUROC$\uparrow$/ FPR95$\downarrow$ & AUROC$\uparrow$/ FPR95$\downarrow$ & AUROC$\uparrow$/ FPR95$\downarrow$ \\
\midrule

Gaussian Noise & 58.4 / 88.8& 68.7 / 76.8& 80.6 / 55.2& 86.1 / 42.4& 90.6 / 29.8 & 76.9 / 58.6\\
Shot Noise & 54.9 / 91.8& 59.8 / 87.1& 73.4 / 69.2& 79.7 / 58.6& 88.5 / 38.5 & 71.3 / 69.0\\
Speckle Noise & 55.0 / 91.8& 63.7 / 83.7& 69.4 / 78.1& 80.5 / 63.2& 89.2 / 45.0 & 71.6 / 72.3\\
Impulse Noise & 66.4 / 81.3& 80.3 / 60.0& 89.4 / 37.4& 97.6 / 9.8& 99.5 / 1.8 & 86.7 / 38.1\\
Defocus Blur & 45.1 / 96.1& 38.5 / 96.7& 33.8 / 97.4& 31.2 / 97.7& 27.7 / 97.9 & 35.2 / 97.2\\
Gaussian Blur & 45.5 / 95.9& 34.0 / 97.4& 30.7 / 97.7& 28.2 / 98.0& 25.7 / 98.1 & 32.8 / 97.4\\
Glass Blur & 57.2 / 93.0 & 55.10 / 94.1 & 46.11 / 95.6 & 60.0 / 92.7 & 50.2 / 94.6 & 53.7 / 94.0 \\
Motion Blur & 38.1 / 96.7& 33.6 / 97.4& 30.8 / 97.7& 30.8 / 97.7& 28.8 / 97.8 & 32.4 / 97.5\\
Zoom Blur & 35.5 / 97.2& 33.2 / 97.4& 31.5 / 97.5& 30.0 / 97.6& 28.2 / 97.7  & 31.7 / 97.5\\
Snow & 68.0 / 79.4& 94.2 / 21.2& 96.7 / 12.9& 99.8 / 0.8& 100.0 / 0.0 & 91.7 / 22.9 \\
Fog & 38.9 / 97.0& 43.8 / 94.9& 49.9 / 92.7& 52.7 / 90.3& 55.2 / 86.8 & 48.1 / 92.2\\
Brightness & 62.6 / 86.2& 85.6 / 47.9& 97.8 / 8.7& 99.8 / 0.8& 100.0 / 0.0 & 89.2 / 28.7 \\
Contrast & 34.1 / 97.7& 38.3 / 96.7& 43.5 / 96.3& 48.3 / 95.7& 50.3 / 95.3 & 42.9 / 96.3\\
Elastic Transform & 42.2 / 96.4& 39.9 / 96.5& 36.1 / 97.1& 36.1 / 96.9& 38.0 / 96.7 & 38.5 / 96.7\\
Pixelate & 48.7 / 95.2& 48.0 / 95.2& 47.8 / 95.6& 47.5 / 95.7& 45.1 / 95.8 & 47.4 / 95.5\\
JPEG Compression & 49.7 / 95.2& 48.7 / 95.5& 48.5 / 95.4& 48.0 / 95.6 & 47.7 / 95.3 & 48.5 / 95.4 \\
Spatter & 55.2 / 91.4& 63.2 / 82.5& 74.8 / 66.1& 66.9 / 84.3& 75.7 / 77.0 & 67.2 / 80.0\\
Saturate & 65.0 / 80.5& 72.6 / 70.6& 63.3 / 88.1& 90.7 / 41.8& 98.9 / 5.3 & 78.1 / 57.3\\
Frost & 97.9 / 7.8& 100.0 / 0.2& 100.0 / 0.1& 99.9 / 0.3& 99.7 / 1.2 & 99.5 / 1.9\\
\bottomrule
\end{tabular}}

\caption{The performance of the AVAE model in detecting out-of-distribution (OOD) covariate shift between CIFAR10 and CIFAR10-C datasets is evaluated. The model is evaluated using \textbf{MSE + KL-divergence + Adversarial} as metric. The model achieves a mean Area Under the Receiver Operating Characteristic (AUROC) of 60.2\% and a False Positive Rate at 95\% True Positive Rate (FPR95) of 73.1\%.}\label{tab:avae_results}}
\end{table*}

\begin{table*}[htbp]
\centering
{\resizebox{1.0\linewidth}{!}{
\begin{tabular}{r|ccccccc}
Severity & 1 & 2 & 3 & 4 & 5 & Average \\
Metric & AUROC$\uparrow$/ FPR95$\downarrow$ & AUROC$\uparrow$/ FPR95$\downarrow$ & AUROC$\uparrow$/ FPR95$\downarrow$ & AUROC$\uparrow$/ FPR95$\downarrow$ & AUROC$\uparrow$/ FPR95$\downarrow$ & AUROC$\uparrow$/ FPR95$\downarrow$ \\
\midrule
Gaussian Noise & 11.78 / 99.4 & 11.76 / 99.06 & 10.31 / 98.96 & 10.12 / 98.9 & 10.32 / 98.98 & 10.86 / 99.06 \\
Shot Noise & 13.26 / 99.62 & 10.25 / 99.54 & 12.32 / 98.8 & 21.2 / 96.8 & 80.4 / 34.66 & 27.49 / 85.88 \\
Speckle Noise & 14.19 / 99.54 & 8.76 / 99.66 & 7.42 / 99.74 & 5.68 / 99.82 & 5.34 / 99.82 & 8.28 / 99.72 \\
Impulse Noise & 82.32 / 54.02 & 91.32 / 26.3 & 94.41 / 15.58 & 96.6 / 8.44 & 97.25 / 6.08 & 92.38 / 22.08 \\
Defocus Blur & 54.86 / 91.56 & 60.7 / 88.62 & 66.93 / 82.18 & 70.74 / 75.0 & 78.47 / 58.0 & 66.34 / 79.07 \\
Gaussian Blur & 55.4 / 91.26 & 66.78 / 82.36 & 70.56 / 75.56 & 73.89 / 69.8 & 78.93 / 56.84 & 69.11 / 75.16 \\
Glass Blur & 29.79 / 96.54 & 28.62 / 97.42 & 22.86 / 98.56 & 46.11 / 89.8 & 26.81 / 97.32 & 30.84 / 95.93 \\
Motion Blur & 58.69 / 90.22 & 63.21 / 86.7 & 67.24 / 82.36 & 66.73 / 84.4 & 69.38 / 81.08 & 65.05 / 84.95 \\
Zoom Blur & 61.45 / 87.6 & 65.43 / 82.92 & 67.54 / 79.86 & 69.11 / 77.4 & 71.03 / 74.16 & 66.91 / 80.39 \\
Snow & 53.74 / 92.08 & 50.7 / 93.08 & 56.26 / 89.42 & 56.2 / 89.48 & 56.5 / 91.08 & 54.68 / 91.03 \\
Fog & 51.45 / 91.84 & 57.41 / 85.8 & 61.76 / 79.46 & 63.38 / 75.6 & 63.97 / 70.48 & 59.6 / 80.64 \\
Brightness & 49.75 / 94.9 & 51.09 / 94.5 & 53.06 / 93.46 & 56.62 / 91.64 & 63.58 / 87.32 & 54.82 / 92.36 \\
Contrast & 59.6 / 86.84 & 75.93 / 62.96 & 81.8 / 49.1 & 88.51 / 33.06 & 96.23 / 8.96 & 80.41 / 48.18 \\
Elastic Transform & 58.2 / 89.46 & 68.92 / 77.98 & 66.32 / 79.52 & 57.31 / 85.66 & 45.73 / 91.22 & 59.3 / 84.77 \\
Pixelate & 88.99 / 25.18 & 96.71 / 5.32 & 97.78 / 3.4 & 99.0 / 1.36 & 99.96 / 0.16 & 96.49 / 7.08 \\
JPEG Compression & 59.64 / 92.2 & 70.8 / 82.96 & 74.59 / 77.98 & 78.17 / 68.92 & 83.57 / 56.64 & 73.35 / 75.74 \\
Spatter & 54.71 / 91.9 & 60.38 / 88.06 & 60.86 / 85.94 & 61.72 / 86.26 & 67.53 / 82.96 & 61.04 / 87.02 \\
Saturate & 72.83 / 78.02 & 94.09 / 15.04 & 53.79 / 94.44 & 72.87 / 78.72 & 83.32 / 52.06 & 75.38 / 63.66 \\
Frost & 37.52 / 97.7 & 38.83 / 97.06 & 35.33 / 96.78 & 33.43 / 96.82 & 30.43 / 96.08 & 35.11 / 96.89 \\
\bottomrule
\end{tabular}}
\caption{The performance of the VAE FRL model in detecting out-of-distribution (OOD) covariate shift between CIFAR10 and CIFAR10-C datasets is evaluated. The model is evaluated using \textbf{Cross Entroy + KL-divergence - Input Complexity} as metric. The model achieves a mean Area Under the Receiver Operating Characteristic (AUROC) of 57.2\% and a False Positive Rate at 95\% True Positive Rate (FPR95) of 76.3\%.}\label{tab:vae_frl_results}}
\end{table*}

\begin{table*}[htbp]
\centering
{\resizebox{1.0\linewidth}{!}{
\begin{tabular}{r|cccccc}
Severity & 1 & 2 & 3 & 4 & 5 & Average \\
Metric & AUROC$\uparrow$/ FPR95$\downarrow$ & AUROC$\uparrow$/ FPR95$\downarrow$ & AUROC$\uparrow$/ FPR95$\downarrow$ & AUROC$\uparrow$/ FPR95$\downarrow$ & AUROC$\uparrow$/ FPR95$\downarrow$ & AUROC$\uparrow$/ FPR95$\downarrow$ \\
\midrule
Gaussian Noise & 100.0 / 0.0 & 100.0 / 0.0 & 100.0 / 0.0 & 100.0 / 0.0 & 100.0 / 0.0 & 100.0 / 0.0 \\
Shot Noise & 99.9 / 0.0 & 99.96 / 0.0 & 99.99 / 0.0 & 100.0 / 0.0 & 100.0 / 0.0 & 99.97 / 0.0 \\
Speckle Noise & 99.73 / 0.1 & 99.86 / 0.0 & 99.89 / 0.0 & 99.93 / 0.0 & 99.95 / 0.0 & 99.87 / 0.02 \\
Impulse Noise & 99.46 / 2.4 & 100.0 / 0.0 & 100.0 / 0.0 & 100.0 / 0.0 & 100.0 / 0.0 & 99.89 / 0.48 \\
Defocus Blur & 44.17 / 95.7 & 31.83 / 97.0 & 20.95 / 98.1 & 17.69 / 98.6 & 9.31 / 99.0  & 24.79 / 97.68 \\
Gaussian Blur & 44.32 / 95.7 & 21.34 / 98.0 & 14.15 / 98.6 & 9.75 / 99.0 & 5.21 / 99.3 & 18.95 / 98.12 \\
Glass Blur & 87.80 / 65.4 & 84.70 / 77.3 & 79.33 / 83.9 & 87.56 / 69.9 & 82.84 / 78.1 & 84.45 / 74.92 \\
Motion Blur & 34.23 / 96.5 & 26.82 / 97.1 & 21.77 / 97.7 & 21.72 / 97.5 & 18.18 / 98.4 & 24.54 / 97.44 \\
Zoom Blur & 27.71 / 97.1 & 21.09 / 97.8 & 17.25 / 98.6 & 14.45 / 98.6 & 11.53 / 98.6 & 18.41 / 98.14 \\
Snow & 62.99 / 88.6 & 74.97 / 80.0 & 71.98 / 81.7 & 70.85 / 87.0 & 71.06 / 90.9 & 70.37 / 85.64 \\
Fog & 44.69 / 95.0 & 33.72 / 96.1 & 27.79 / 96.6 & 24.17 / 96.5 & 22.13 / 95.2 & 30.5 / 95.88 \\
Brightness & 57.86 / 93.6 & 63.6 / 91.3 & 67.67 / 89.8 & 71.17 / 88.7 & 73.61 / 88.2 & 66.78 / 90.32 \\
Contrast & 39.17 / 95.6 & 20.93 / 98.6 & 14.86 / 98.7 & 8.79 / 99.3 & 2.23 / 99.8 & 17.2 / 98.4 \\
Elastic Transform & 38.87 / 95.8 & 34.2 / 96.8 & 27.75 / 97.1 & 36.17 / 96.5 & 51.48 / 93.9 & 37.69 / 96.02 \\
Pixelate & 57.09 / 93.6 & 62.0 / 92.3 & 63.52 / 91.4 & 67.16 / 90.3 & 67.01 / 90.8 & 63.36 / 91.68 \\
JPEG Compression & 49.32 / 90.9 & 44.23 / 93.8 & 42.04 / 94.5 & 39.94 / 95.7 & 36.21 / 96.2 & 42.35 / 94.22 \\
Spatter & 68.03 / 84.3 & 81.11 / 64.5 & 88.64 / 39.6 & 75.23 / 70.5 & 88.36 / 39.6 & 80.27 / 59.7 \\
Saturate & 23.9 / 97.4 & 12.14 / 98.7 & 69.4 / 88.2 & 88.04 / 52.1 & 92.2 / 42.8 & 57.14 / 75.84 \\
Frost & 73.96 / 78.0 & 79.53 / 79.0 & 82.89 / 66.1 & 83.17 / 60.2 & 84.26 / 48.2  & 80.76 / 66.3 \\
\bottomrule
\end{tabular}}

\caption{The performance of the GLOW model in detecting out-of-distribution (OOD) covariate shift between CIFAR10 and CIFAR10-C datasets is evaluated. The model is evaluated using \textbf{log-likelihood} as metric. The model achieves a mean Area Under the Receiver Operating Characteristic (AUROC) of 58.8\% and a False Positive Rate at 95\% True Positive Rate (FPR95) of 69.5\%.\label{tab:glow_results_ll}}}
\end{table*}

\begin{table*}[htbp]
\centering
{\resizebox{1.0\linewidth}{!}{
\begin{tabular}{r|ccccccc}
Severity & 1 & 2 & 3 & 4 & 5 & Average \\
Metric & AUROC$\uparrow$/ FPR95$\downarrow$ & AUROC$\uparrow$/ FPR95$\downarrow$ & AUROC$\uparrow$/ FPR95$\downarrow$ & AUROC$\uparrow$/ FPR95$\downarrow$ & AUROC$\uparrow$/ FPR95$\downarrow$ & AUROC$\uparrow$/ FPR95$\downarrow$ \\
\midrule
Gaussian Noise & 0.01 / 100.0 & 0.01 / 100.0 & 0.21 / 100.0 & 0.18 / 100.0 & 0.47 / 100.0  & 0.18 / 100.0 \\
Shot Noise & 0.45 / 100.0 & 0.39 / 100.0 & 0.33 / 100.0 & 0.39 / 100.0 & 0.73 / 100.0  & 0.46 / 100.0 \\
Speckle Noise & 0.74 / 100.0 & 0.61 / 100.0 & 0.58 / 100.0 & 0.79 / 100.0 & 1.77 / 100.0  & 0.9 / 100.0 \\
Impulse Noise & 13.31 / 99.1 & 10.99 / 100.0 & 12.43 / 100.0 & 19.25 / 100.0 & 29.25 / 99.8  & 17.05 / 99.78 \\
Defocus Blur & 55.71 / 92.0 & 60.23 / 83.9 & 65.64 / 73.1 & 72.44 / 59.5 & 80.08 / 45.2  & 66.82 / 70.74 \\
Gaussian Blur & 55.44 / 92.9 & 65.11 / 74.9 & 71.07 / 62.5 & 76.51 / 53.4 & 85.53 / 36.6  & 70.73 / 64.06 \\
Glass Blur & 14.78 / 100.0 & 17.24 / 99.8 & 19.79 / 99.0 & 14.75 / 100.0 & 16.81 / 99.5  & 16.67 / 99.66 \\
Motion Blur & 59.17 / 86.5 & 62.82 / 80.7 & 66.05 / 78.2 & 66.08 / 76.0 & 68.87 / 71.4  & 64.6 / 78.56 \\
Zoom Blur & 62.07 / 80.0 & 66.15 / 72.1 & 69.17 / 66.9 & 71.99 / 61.1 & 75.62 / 54.7  & 69.0 / 66.96 \\
Snow & 46.25 / 94.8 & 40.12 / 96.3 & 40.56 / 96.2 & 39.72 / 95.8 & 39.95 / 96.2  & 41.32 / 95.86 \\
Fog & 58.87 / 92.4 & 62.46 / 88.3 & 64.97 / 83.6 & 66.07 / 80.0 & 69.08 / 74.1  & 64.29 / 83.68 \\
Brightness & 49.3 / 95.8 & 44.74 / 96.7 & 39.64 / 97.0 & 35.2 / 97.7 & 29.03 / 98.6  & 39.58 / 97.16 \\
Contrast & 60.38 / 90.6 & 66.07 / 81.4 & 71.51 / 73.0 & 77.7 / 62.3 & 91.2 / 30.3  & 73.37 / 67.52 \\
Elastic Transform & 53.64 / 89.7 & 55.91 / 86.1 & 59.33 / 79.9 & 48.9 / 86.7 & 34.59 / 94.0  & 50.47 / 87.28 \\
Pixelate & 41.61 / 96.1 & 36.16 / 96.8 & 33.18 / 96.8 & 29.0 / 97.3 & 26.97 / 97.3  & 33.38 / 96.86 \\
JPEG Compression & 83.05 / 54.0 & 84.49 / 47.8 & 85.03 / 47.2 & 84.93 / 44.4 & 84.12 / 47.0  & 84.32 / 48.08 \\
Spatter & 33.34 / 96.8 & 20.16 / 97.6 & 12.59 / 98.0 & 26.44 / 97.3 & 13.98 / 98.3  & 21.3 / 97.6 \\
Saturate & 71.75 / 68.0 & 92.63 / 24.3 & 45.27 / 97.6 & 17.6 / 99.8 & 9.36 / 100.0  & 47.32 / 77.94 \\
Frost & 36.11 / 97.3 & 29.7 / 98.7 & 25.6 / 99.8 & 26.3 / 99.5 & 25.86 / 99.7  & 28.71 / 99.0 \\
\bottomrule
\end{tabular}}
\caption{The performance of the GLOW model in detecting out-of-distribution (OOD) covariate shift between CIFAR10 and CIFAR10-C datasets is evaluated. The model is evaluated using \textbf{typicality} as metric. The model achieves a mean Area Under the Receiver Operating Characteristic (AUROC) of 41.6\% and a False Positive Rate at 95\% True Positive Rate (FPR95) of 85.8\%.\label{tab:glow_results_typ}}}
\end{table*}

\begin{table*}[htbp]
\centering
{\resizebox{1.0\linewidth}{!}{
\begin{tabular}{r|ccccccc}
Severity & 1 & 2 & 3 & 4 & 5 & Average \\
Metric & AUROC$\uparrow$/ FPR95$\downarrow$ & AUROC$\uparrow$/ FPR95$\downarrow$ & AUROC$\uparrow$/ FPR95$\downarrow$ & AUROC$\uparrow$/ FPR95$\downarrow$ & AUROC$\uparrow$/ FPR95$\downarrow$ & AUROC$\uparrow$/ FPR95$\downarrow$\\
\midrule
Gaussian Noise & 99.99 / 0.1 & 100.0 / 0.0 & 100.0 / 0.0 & 100.0 / 0.0 & 100.0 / 0.0  & 100.0 / 0.02 \\
Shot Noise & 99.62 / 0.5 & 99.82 / 0.1 & 99.94 / 0.0 & 99.96 / 0.0 & 99.96 / 0.0  & 99.86 / 0.12 \\
Speckle Noise & 99.39 / 1.5 & 99.76 / 0.3 & 99.81 / 0.1 & 99.85 / 0.1 & 99.89 / 0.0  & 99.74 / 0.4 \\
Impulse Noise & 95.95 / 13.1 & 99.77 / 1.0 & 99.98 / 0.1 & 100.0 / 0.0 & 100.0 / 0.0  & 99.14 / 2.84 \\
Defocus Blur & 48.67 / 94.8 & 47.99 / 95.7 & 53.14 / 88.1 & 57.01 / 92.1 & 71.64 / 73.1  & 55.69 / 88.76 \\
Gaussian Blur & 48.74 / 95.2 & 52.7 / 88.6 & 60.77 / 81.1 & 69.22 / 72.4 & 81.95 / 44.6  & 62.68 / 76.38 \\
Glass Blur & 83.37 / 68.1 & 80.56 / 73.3 & 75.60 / 78.3 & 83.52 / 66.2 & 79.06 / 74.8  & 80.42 / 72.14 \\
Motion Blur & 46.99 / 95.8 & 49.89 / 94.2 & 53.37 / 89.9 & 53.72 / 89.9 & 57.48 / 87.1  & 52.29 / 91.38 \\
Zoom Blur & 48.54 / 92.9 & 52.78 / 89.9 & 56.63 / 86.0 & 60.52 / 84.1 & 66.06 / 79.1  & 56.91 / 86.4 \\
Snow & 51.08 / 92.5 & 60.15 / 82.8 & 57.14 / 87.0 & 58.05 / 89.9 & 60.55 / 85.5  & 57.39 / 87.54 \\
Fog & 49.6 / 94.2 & 50.45 / 94.8 & 52.6 / 93.4 & 53.88 / 88.3 & 54.22 / 90.5  & 52.15 / 92.24 \\
Brightness & 53.11 / 91.8 & 56.2 / 89.7 & 59.07 / 89.6 & 62.44 / 89.6 & 67.19 / 92.5  & 59.6 / 90.64 \\
Contrast & 49.61 / 94.3 & 57.67 / 84.6 & 64.43 / 77.4 & 73.96 / 65.3 & 90.95 / 29.6  & 67.32 / 70.24 \\
Elastic Transform & 46.48 / 94.9 & 46.47 / 94.3 & 47.27 / 93.5 & 43.97 / 92.8 & 50.95 / 90.2  & 47.03 / 93.14 \\
Pixelate & 51.88 / 95.4 & 55.49 / 94.3 & 57.67 / 94.0 & 61.57 / 92.1 & 62.39 / 91.2  & 57.8 / 93.4 \\
JPEG Compression & 57.41 / 65.7 & 57.93 / 66.3 & 58.43 / 68.1 & 58.41 / 72.2 & 58.23 / 75.6  & 58.08 / 69.58 \\
Spatter & 54.38 / 95.9 & 70.71 / 81.6 & 81.41 / 58.0 & 60.72 / 92.2 & 78.86 / 64.6  & 69.22 / 78.46 \\
Saturate & 54.15 / 94.0 & 80.13 / 62.6 & 60.02 / 82.5 & 79.14 / 75.6 & 87.57 / 58.7  & 72.2 / 74.68 \\
Frost & 60.71 / 86.2 & 69.18 / 80.3 & 71.64 / 78.7 & 70.37 / 78.7 & 69.45 / 77.3  & 68.27 / 80.24 \\
\bottomrule
\end{tabular}}
\caption{The performance of the GLOW model in detecting out-of-distribution (OOD) covariate shift between CIFAR10 and CIFAR10-C datasets is evaluated. The model is evaluated using the \textbf{normalized score distance} as metric. The model achieves a mean Area Under the Receiver Operating Characteristic (AUROC) of 69.25\% and a False Positive Rate at 95\% True Positive Rate (FPR95) of 65.57\%.\label{tab:glow_results_nsh}}}
\end{table*}

\begin{table*}[htbp]
\centering
{\resizebox{1.0\linewidth}{!}{
\begin{tabular}{r|cccccc}
Severity & 1 & 2 & 3 & 4 & 5 & Average \\
Metric & AUROC$\uparrow$/ FPR95$\downarrow$ & AUROC$\uparrow$/ FPR95$\downarrow$ & AUROC$\uparrow$/ FPR95$\downarrow$ & AUROC$\uparrow$/ FPR95$\downarrow$ & AUROC$\uparrow$/ FPR95$\downarrow$ & AUROC$\uparrow$/ FPR95$\downarrow$\\
\midrule
Gaussian Noise & 100.0 / 0.0 & 100.0 / 0.0 & 100.0 / 0.0 & 100.0 / 0.0 & 100.0 / 0.0  & 100.0 / 0.0 \\
Shot Noise & 99.97 / 0.03 & 100.0 / 0.0 & 100.0 / 0.0 & 100.0 / 0.0 & 100.0 / 0.0  & 99.99 / 0.01 \\
Speckle Noise & 99.87 / 0.46 & 99.98 / 0.0 & 99.99 / 0.0 & 100.0 / 0.0 & 100.0 / 0.0  & 99.97 / 0.09 \\
Impulse Noise & 100.0 / 0.0 & 100.0 / 0.0 & 100.0 / 0.0 & 100.0 / 0.0 & 100.0 / 0.0  & 100.0 / 0.0 \\
Defocus Blur & 41.76 / 96.56 & 26.48 / 98.66 & 15.67 / 99.39 & 12.81 / 99.49 & 6.67 / 99.73  & 20.68 / 98.77 \\
Gaussian Blur & 42.08 / 96.55 & 16.35 / 99.35 & 10.53 / 99.64 & 7.43 / 99.72 & 4.36 / 99.79  & 16.15 / 99.01 \\
Glass Blur & 94.10 / 26.6 & 92.71 / 32.9 & 85.63 / 53.3 & 95.30 / 22.3 & 90.04 / 41.4  & 91.56 / 35.3 \\
Motion Blur & 28.51 / 98.22 & 20.43 / 99.04 & 15.41 / 99.35 & 15.39 / 99.34 & 12.06 / 99.52  & 18.36 / 99.09 \\
Zoom Blur & 21.57 / 99.06 & 15.85 / 99.3 & 12.66 / 99.47 & 10.36 / 99.54 & 8.13 / 99.67  & 13.71 / 99.41 \\
Snow & 66.81 / 76.76 & 78.15 / 57.68 & 75.76 / 62.43 & 74.43 / 66.1 & 77.02 / 62.06  & 74.43 / 65.01 \\
Fog & 37.97 / 96.31 & 22.81 / 98.66 & 15.71 / 99.16 & 11.1 / 99.39 & 7.29 / 99.4  & 18.98 / 98.58 \\
Brightness & 56.99 / 91.95 & 63.15 / 88.38 & 68.39 / 84.34 & 73.15 / 78.64 & 78.57 / 72.78  & 68.05 / 83.22 \\
Contrast & 31.23 / 97.77 & 11.0 / 99.63 & 6.3 / 99.73 & 2.84 / 99.84 & 0.51 / 99.98  & 10.38 / 99.39 \\
Elastic Transform & 34.78 / 97.95 & 29.44 / 98.41 & 22.61 / 98.92 & 29.91 / 98.15 & 46.57 / 94.68  & 32.66 / 97.62 \\
Pixelate & 58.57 / 91.1 & 63.7 / 87.87 & 65.79 / 86.0 & 70.44 / 81.05 & 74.14 / 75.92  & 66.53 / 84.39 \\
JPEG Compression & 52.06 / 90.7 & 48.68 / 92.9 & 48.04 / 93.33 & 46.66 / 94.63 & 44.78 / 95.83  & 48.04 / 93.48 \\
Spatter & 77.96 / 56.4 & 90.66 / 25.05 & 92.64 / 18.55 & 90.29 / 33.44 & 97.05 / 11.91  & 89.72 / 29.07 \\
Saturate & 22.04 / 98.92 & 16.43 / 99.45 & 71.69 / 79.1 & 91.61 / 33.98 & 96.68 / 13.23  & 59.69 / 64.94 \\
Frost & 70.48 / 72.15 & 79.48 / 58.54 & 82.12 / 52.39 & 79.68 / 56.79 & 78.54 / 57.5  & 78.06 / 59.47 \\
\bottomrule
\end{tabular}}

\caption{The performance of the CovariateFlow model in detecting out-of-distribution (OOD) covariate shift between CIFAR10 and CIFAR10-C datasets is evaluated. The model is evaluated using \textbf{log-likelihood} as metric. The model achieves a mean Area Under the Receiver Operating Characteristic (AUROC) of 58.3\% and a False Positive Rate at 95\% True Positive Rate (FPR95) of 63.5\%.\label{tab:covariate_results_ll}}}
\end{table*}

\begin{table*}[htbp]
\centering
{\resizebox{1.0\linewidth}{!}{
\begin{tabular}{r|ccccccc}
Severity & 1 & 2 & 3 & 4 & 5 & Average \\
Metric & AUROC$\uparrow$/ FPR95$\downarrow$ & AUROC$\uparrow$/ FPR95$\downarrow$ & AUROC$\uparrow$/ FPR95$\downarrow$ & AUROC$\uparrow$/ FPR95$\downarrow$ & AUROC$\uparrow$/ FPR95$\downarrow$ & AUROC$\uparrow$/ FPR95$\downarrow$\\
\midrule
Gaussian Noise & 6.99 / 100.0 & 1.75 / 100.0 & 0.44 / 100.0 & 0.19 / 100.0 & 0.08 / 100.0  & 1.89 / 100.0 \\
Shot Noise & 13.91 / 99.99 & 7.24 / 100.0 & 1.54 / 100.0 & 0.86 / 100.0 & 0.36 / 100.0  & 4.78 / 100.0 \\
Speckle Noise & 14.62 / 99.99 & 5.91 / 100.0 & 3.92 / 100.0 & 1.79 / 100.0 & 0.88 / 100.0  & 5.42 / 100.0 \\
Impulse Noise & 1.72 / 100.0 & 0.25 / 100.0 & 0.09 / 100.0 & 0.03 / 100.0 & 0.03 / 100.0  & 0.42 / 100.0 \\
Defocus Blur & 56.28 / 91.03 & 67.29 / 77.35 & 75.52 / 58.82 & 81.35 / 48.12 & 87.62 / 33.84  & 73.61 / 61.83 \\
Gaussian Blur & 56.14 / 91.32 & 75.85 / 58.06 & 81.0 / 44.39 & 84.6 / 36.37 & 89.52 / 27.93  & 77.42 / 51.61 \\
Glass Blur & 31.48 / 99.64 & 33.31 / 99.53 & 40.55 / 98.72 & 27.37 / 99.75 & 35.57 / 99.28  & 33.66 / 99.38 \\
Motion Blur & 66.94 / 78.89 & 73.39 / 67.27 & 77.53 / 57.47 & 77.39 / 56.9 & 80.2 / 50.71  & 75.09 / 62.25 \\
Zoom Blur & 75.41 / 64.55 & 77.96 / 56.4 & 81.04 / 48.43 & 82.35 / 44.68 & 84.65 / 38.59  & 80.28 / 50.53 \\
Snow & 40.84 / 97.16 & 34.39 / 98.25 & 34.76 / 98.08 & 35.47 / 97.95 & 36.02 / 98.05  & 36.3 / 97.9 \\
Fog & 59.09 / 88.58 & 69.74 / 74.13 & 74.34 / 62.76 & 77.76 / 54.6 & 81.51 / 45.02  & 72.49 / 65.02 \\
Brightness & 46.11 / 96.49 & 42.79 / 97.53 & 39.77 / 98.37 & 36.64 / 98.91 & 32.23 / 99.42  & 39.51 / 98.14 \\
Contrast & 63.27 / 83.41 & 77.58 / 55.23 & 81.85 / 43.87 & 85.81 / 34.14 & 91.13 / 23.89  & 79.93 / 48.11 \\
Elastic Transform & 62.73 / 85.85 & 66.2 / 80.71 & 71.06 / 71.71 & 65.88 / 79.12 & 55.36 / 90.91  & 64.25 / 81.66 \\
Pixelate & 49.57 / 96.24 & 50.82 / 96.96 & 50.71 / 97.01 & 53.34 / 97.24 & 58.73 / 96.77  & 52.63 / 96.84 \\
JPEG Compression & 53.62 / 93.49 & 56.54 / 92.86 & 58.53 / 92.25 & 59.89 / 91.32 & 61.29 / 90.17  & 57.97 / 92.02 \\
Spatter & 40.06 / 97.67 & 30.67 / 98.94 & 27.04 / 99.36 & 24.72 / 99.29 & 16.57 / 99.89  & 27.81 / 99.03 \\
Saturate & 64.44 / 77.58 & 66.78 / 72.5 & 36.82 / 99.02 & 21.56 / 99.89 & 13.69 / 100.0  & 40.66 / 89.8 \\
Frost & 41.35 / 97.29 & 38.0 / 98.22 & 39.38 / 98.04 & 41.29 / 97.43 & 43.61 / 96.68  & 40.73 / 97.53 \\
\bottomrule
\end{tabular}}
\caption{The performance of the CovariateFlow model in detecting out-of-distribution (OOD) covariate shift between CIFAR10 and CIFAR10-C datasets is evaluated. The model is evaluated using \textbf{typicality} as metric. The model achieves a mean Area Under the Receiver Operating Characteristic (AUROC) of 45.5\% and a False Positive Rate at 95\% True Positive Rate (FPR95) of 83.8\%.\label{tab:covariate_results_typ}}}
\end{table*}

\begin{table*}[htbp]
\centering
{\resizebox{1.0\linewidth}{!}{
\begin{tabular}{r|ccccccc}
Severity & 1 & 2 & 3 & 4 & 5 & Average \\
Metric & AUROC$\uparrow$/ FPR95$\downarrow$ & AUROC$\uparrow$/ FPR95$\downarrow$ & AUROC$\uparrow$/ FPR95$\downarrow$ & AUROC$\uparrow$/ FPR95$\downarrow$ & AUROC$\uparrow$/ FPR95$\downarrow$ & AUROC$\uparrow$/ FPR95$\downarrow$\\
\midrule
Gaussian Noise & 99.46 / 0.63 & 99.65 / 0.4 & 99.79 / 0.25 & 99.81 / 0.2 & 99.82 / 0.19  & 99.71 / 0.33 \\
Shot Noise & 99.19 / 1.16 & 99.46 / 0.71 & 99.69 / 0.43 & 99.76 / 0.39 & 99.81 / 0.25  & 99.58 / 0.59 \\
Speckle Noise & 98.96 / 2.07 & 99.48 / 0.78 & 99.59 / 0.68 & 99.72 / 0.47 & 99.79 / 0.39  & 99.51 / 0.88 \\
Impulse Noise & 99.68 / 0.5 & 99.84 / 0.19 & 99.88 / 0.15 & 99.91 / 0.12 & 99.92 / 0.09  & 99.85 / 0.21 \\
Defocus Blur & 50.33 / 94.96 & 58.69 / 92.74 & 70.05 / 84.46 & 75.37 / 78.97 & 85.71 / 52.89  & 68.03 / 80.8 \\
Gaussian Blur & 50.14 / 95.3 & 69.37 / 85.8 & 77.29 / 73.98 & 82.67 / 61.39 & 89.41 / 36.32  & 73.78 / 70.56 \\
Glass Blur & 89.41 / 45.08 & 87.57 / 53.53 & 77.47 / 74.20 & 91.13 / 38.94 & 83.31 / 64.63  & 85.78 / 55.28 \\
Motion Blur & 57.66 / 93.61 & 65.53 / 89.88 & 72.16 / 83.91 & 72.05 / 84.16 & 76.56 / 75.4  & 68.79 / 85.39 \\
Zoom Blur & 64.69 / 91.29 & 70.9 / 85.13 & 75.56 / 78.26 & 78.72 / 71.08 & 82.34 / 63.09  & 74.44 / 77.77 \\
Snow & 50.58 / 95.61 & 61.29 / 91.65 & 58.24 / 93.14 & 57.31 / 93.96 & 60.98 / 92.24  & 57.68 / 93.32 \\
Fog & 51.21 / 94.85 & 62.74 / 90.19 & 70.43 / 83.68 & 76.32 / 74.28 & 81.99 / 56.85  & 68.54 / 79.97 \\
Brightness & 50.58 / 95.27 & 53.04 / 95.21 & 56.92 / 94.54 & 61.02 / 93.45 & 68.04 / 91.46  & 57.92 / 93.99 \\
Contrast & 55.32 / 93.67 & 76.67 / 75.78 & 83.86 / 54.76 & 90.12 / 29.31 & 95.89 / 10.34  & 80.37 / 52.77 \\
Elastic Transform & 53.65 / 94.88 & 56.87 / 94.16 & 62.39 / 91.69 & 55.03 / 94.55 & 47.27 / 96.18  & 55.04 / 94.29 \\
Pixelate & 52.09 / 95.21 & 54.97 / 95.39 & 56.66 / 95.11 & 61.11 / 94.28 & 66.07 / 92.41  & 58.18 / 94.48 \\
JPEG Compression & 46.36 / 96.46 & 46.7 / 96.45 & 47.28 / 96.54 & 47.75 / 96.6 & 48.96 / 96.57  & 47.41 / 96.52 \\
Spatter & 60.21 / 93.94 & 81.88 / 52.79 & 85.73 / 38.68 & 80.65 / 71.55 & 93.5 / 23.28  & 80.39 / 56.05 \\
Saturate & 60.09 / 89.53 & 66.52 / 84.43 & 59.65 / 94.4 & 85.27 / 62.34 & 93.93 / 24.4  & 73.09 / 71.02 \\
Frost & 53.51 / 95.45 & 64.2 / 91.5 & 67.86 / 89.29 & 64.31 / 91.19 & 62.17 / 92.49  & 62.41 / 91.98 \\
\bottomrule
\end{tabular}}

\caption{The performance of the CovariateFlow model in detecting out-of-distribution (OOD) covariate shift between CIFAR10 and CIFAR10-C datasets is evaluated. The model is evaluated using \textbf{normalized score distance} as metric. The model achieves a mean Area Under the Receiver Operating Characteristic (AUROC) of 74.9\% and a False Positive Rate at 95\% True Positive Rate (FPR95) of 61.7\%.\label{tab:covariate_results_nsd}}}
\end{table*}

\begin{table}[]
    \centering
    {\resizebox{1.0\linewidth}{!}{
    \begin{tabular}{r|ccccccccccc}
    \toprule
     Model & VAE\hspace{1mm} & AVAE & VAE & DDPM & DDPM & GLOW & GLOW & GLOW & CovFlow & CovFlow & CovFlow \\
    Evaluation & ALL & ALL & FLR & T150 & T20 & LL & Typ & NSD & LL & Typ & NSD \\
    \midrule
    Gaussian Noise & 85.7 & 76.9 & 10.9 & 64.6 & 92.8 & 100.0 & 0.2 & 100.0 & 100.0 & 1.9 & 99.7 \\
    Shot Noise & 79.2 & 71.3 & 27.5 & 61.3 & 87.9 & 100.0 & 0.5 & 99.9 & 100.0 & 4.8 & 99.6 \\
    Speckle Noise & 79.3 & 71.6 & 8.2 & 61.3 & 85.3 & 99.9 & 0.9 & 99.7 & 100.0 & 5.4 & 99.5 \\
    Impulse Noise & 92.8 & 86.6 & 92.4 & 77.0 & 97.3 & 99.9 & 17.0 & 99.1 & 100.0 & 0.4 & 99.8 \\
    Defocus Blur & 26.6 & 35.3 & 66.3 & 70.4 & 63.7 & 24.8 & 66.8 & 55.7 & 20.7 & 73.6 & 68.0 \\
    Gaussian Blur & 22.5 & 32.8 & 69.1 & 78.5 & 71.0 & 19.0 & 70.7 & 62.7 & 16.2 & 77.4 & 73.8 \\
    Glass Blur & 55.5 & 53.7 & 30.8 & 64.9 & 73.1 & 84.4 & 16.7 & 80.4 & 91.6 & 33.7 & 85.8 \\
    Motion Blur & 21.6 & 32.4 & 65.1 & 78.6 & 68.7 & 24.5 & 64.6 & 52.3 & 18.4 & 75.1 & 68.8 \\
    Zoom Blur & 21.3 & 31.7 & 66.9 & 80.0 & 69.8 & 18.4 & 69.0 & 56.9 & 13.7 & 80.3 & 74.4 \\
    Snow & 66.1 & 91.7 & 54.7 & 50.1 & 61.2 & 70.4 & 41.3 & 57.4 & 74.4 & 36.3 & 57.7 \\
    Fog & 14.0 & 48.1 & 59.6 & 78.6 & 70.0 & 30.5 & 64.3 & 52.2 & 19.0 & 72.5 & 68.5 \\
    Brightness & 56.1 & 89.2 & 54.8 & 47.4 & 51.5 & 66.8 & 39.6 & 59.6 & 68.0 & 39.5 & 57.9 \\
    Contrast & 6.3 & 42.9 & 80.4 & 50.4 & 47.2 & 17.2 & 73.4 & 67.3 & 10.4 & 79.9 & 80.4 \\
    Elastic Transform & 31.2 & 38.5 & 59.3 & 58.8 & 52.5 & 37.7 & 50.5 & 47.0 & 32.7 & 64.2 & 55.0 \\
    Pixelate & 46.7 & 47.4 & 96.5 & 56.5 & 55.4 & 63.4 & 33.4 & 57.8 & 66.5 & 52.6 & 58.2 \\
    JPEG Compression & 48.0 & 48.5 & 73.3 &57.4 & 56.4 & 42.4 & 84.3 & 58.1 & 48.0 & 58.0 & 47.4 \\
    Spatter & 72.1 & 67.2 & 61.0 & 59.6 & 67.6 & 80.3 & 21.3 & 69.2 & 89.7 & 27.8 & 80.4 \\
    Saturate & 57.9 & 78.1 & 75.3 & 52.2 & 56.6 & 57.1 & 47.3 & 72.2 & 59.7 & 40.7 & 73.1 \\
    Frost & 47.2 & 99.5 & 35.1 & 53.8 & 60.3 & 80.8 & 28.7 & 68.3 & 78.1 & 40.7 & 62.4 \\
    \midrule
    Average & 48.9 & 60.2 & 57.2 &63.1 &67.5 & 57.7 & 41.6 & 69.3 & 58.3 & 45.5 & \textbf{74.9} \\
    \bottomrule
    \end{tabular}}}
    \caption{Comparison of the performance (AUROC) of all the employed models at detecting every CIFAR10(-C) OOD degredation type.}
    \label{tab:models_per_degredation}
\end{table}

\newpage
\subsection{Detailed Results on ImageNet200 vs. ImageNet200-C}\label{sec:supp_detailed_results_tin}
The following section depicts detailed results obtained with various models on our experiments with ID ImageNet200 and ImageNet200-C as OOD. The results are depicted in order of presentation: DDPM T20-LPIPS+MSE~(\ref{tab:ddpm_t20_tin}), GLOW-LL~(Table~\ref{tab:glow_results_ll_tin}), GLOW-Typicality~(Table~\ref{tab:glow_results_typ_tin}), GLOW-NSD~(Table~\ref{tab:glow_results_nsh}), CovariateFlow-LL~(Table~\ref{tab:covariate_results_ll_tin}), CovariateFlow-Typicality~(Table~\ref{tab:covariate_results_typicality_tin}) and CovariateFlow-NSD~(Table~\ref{tab:covariate_results_nsd_tin}).

\begin{table*}[htbp]
\centering
{\resizebox{1.0\linewidth}{!}{
\begin{tabular}{r|cccccc}
Severity & 1 & 2 & 3 & 4 & 5 & Average \\
Metric & AUROC$\uparrow$/ FPR95$\downarrow$ & AUROC$\uparrow$/ FPR95$\downarrow$ & AUROC$\uparrow$/ FPR95$\downarrow$ & AUROC$\uparrow$/ FPR95$\downarrow$ & AUROC$\uparrow$/ FPR95$\downarrow$ & AUROC$\uparrow$/ FPR95$\downarrow$ \\
\midrule
Brightness & 39.0 / 94.3 & 35.45 / 95.0 & 33.15 / 96.3 & 30.04 / 97.2 & 28.51 / 97.3  & 33.23 / 96.02 \\
Contrast & 59.52 / 86.7 & 64.78 / 85.7 & 73.87 / 80.5 & 85.18 / 63.3 & 89.73 / 51.5  & 74.62 / 73.54 \\
Defocus Blur & 57.46 / 91.8 & 67.45 / 87.1 & 82.33 / 72.9 & 96.75 / 16.8 & 98.77 / 3.7  & 80.55 / 54.46 \\
Elastic Transform & 49.89 / 92.1 & 49.37 / 93.8 & 54.55 / 92.1 & 52.63 / 92.1 & 50.93 / 93.6  & 51.47 / 92.74 \\
Fog & 59.22 / 85.7 & 70.98 / 79.8 & 81.35 / 64.7 & 91.6 / 43.0 & 95.54 / 25.1  & 79.74 / 59.66 \\
Frost & 34.42 / 96.2 & 40.97 / 96.2 & 48.61 / 93.0 & 52.6 / 92.1 & 57.38 / 89.6  & 46.8 / 93.42 \\
Gaussian Noise & 34.87 / 96.5 & 74.38 / 63.2 & 91.73 / 28.0 & 96.13 / 14.2 & 97.71 / 6.9  & 78.96 / 41.76 \\
Glass Blur & 56.65 / 87.2 & 49.07 / 92.5 & 50.19 / 93.2 & 59.12 / 90.2 & 82.11 / 72.6  & 59.43 / 87.14 \\
Impulse Noise & 49.4 / 88.2 & 66.25 / 75.1 & 89.9 / 30.0 & 95.79 / 14.9 & 98.0 / 5.4  & 79.87 / 42.72 \\
JPEG Compression & 43.2 / 93.9 & 41.98 / 94.8 & 47.62 / 93.6 & 47.03 / 92.8 & 52.38 / 92.1  & 46.44 / 93.44 \\
Motion Blur & 54.12 / 92.1 & 59.06 / 88.2 & 70.23 / 83.8 & 76.9 / 77.9 & 82.41 / 66.7  & 68.54 / 81.74 \\
Pixelate & 41.02 / 94.6 & 42.57 / 94.7 & 49.44 / 91.9 & 52.49 / 89.9 & 54.62 / 89.3  & 48.03 / 92.08 \\
Shot Noise & 40.79 / 94.2 & 62.07 / 81.1 & 80.72 / 60.5 & 88.17 / 48.3 & 94.85 / 20.5  & 73.32 / 60.92 \\
Snow & 45.58 / 94.6 & 58.46 / 90.2 & 44.76 / 97.2 & 38.28 / 97.3 & 39.68 / 97.2  & 45.35 / 95.3 \\
Zoom Blur & 64.0 / 87.5 & 71.22 / 81.5 & 77.86 / 73.2 & 83.17 / 62.9 & 87.45 / 55.1  & 76.74 / 72.04 \\
\bottomrule
\end{tabular}}
\caption{The performance of the DDPM model in detecting out-of-distribution (OOD) covariate shift between ImageNet200 and ImageNet200-C datasets is evaluated. The model is evaluated using \textbf{T20-LPIPS+MSE} as metric. The model achieves a mean Area Under the Receiver Operating Characteristic (AUROC) of 62.87\% and a False Positive Rate at 95\% True Positive Rate (FPR95) of 75.80\%.\label{tab:ddpm_t20_tin}}}
\end{table*}


\begin{table*}[htbp]
\centering
{\resizebox{1.0\linewidth}{!}{
\begin{tabular}{r|ccccccc}
Severity & 1 & 2 & 3 & 4 & 5 & Average\\
Metric & AUROC$\uparrow$/ FPR95$\downarrow$ & AUROC$\uparrow$/ FPR95$\downarrow$ & AUROC$\uparrow$/ FPR95$\downarrow$ & AUROC$\uparrow$/ FPR95$\downarrow$ & AUROC$\uparrow$/ FPR95$\downarrow$  & AUROC$\uparrow$/ FPR95$\downarrow$ \\
\midrule
Brightness & 38.03 / 96.5 & 42.45 / 95.9 & 45.58 / 95.6 & 47.64 / 95.9 & 48.21 / 96.5  & 44.38 / 96.08 \\
Contrast & 5.8 / 99.8 & 3.0 / 99.8 & 1.13 / 99.9 & 0.15 / 100.0 & 0.01 / 100.0  & 2.02 / 99.9 \\
Defocus Blur & 17.3 / 98.1 & 14.11 / 98.2 & 9.92 / 99.3 & 5.12 / 99.7 & 4.0 / 99.8  & 10.09 / 99.02 \\
Elastic Transform & 23.8 / 98.0 & 22.13 / 98.0 & 18.54 / 98.1 & 19.07 / 98.1 & 21.5 / 97.8  & 21.01 / 98.0 \\
Fog & 18.73 / 98.1 & 12.86 / 99.3 & 9.19 / 99.6 & 5.73 / 99.8 & 4.0 / 99.8  & 10.1 / 99.32 \\
Frost & 41.85 / 93.7 & 43.38 / 92.1 & 42.83 / 91.5 & 43.73 / 90.2 & 44.57 / 88.6  & 43.27 / 91.22 \\
Gaussian Noise & 65.57 / 65.5 & 97.55 / 6.9 & 99.71 / 0.8 & 99.96 / 0.2 & 99.99 / 0.0  & 92.56 / 14.68 \\
Glass Blur & 50.16 / 93.8 & 23.86 / 97.5 & 15.38 / 98.1 & 11.67 / 98.2 & 6.53 / 99.5  & 21.52 / 97.42 \\
Impulse Noise & 69.23 / 64.7 & 94.31 / 16.7 & 99.73 / 0.5 & 99.98 / 0.1 & 100.0 / 0.0  & 92.65 / 16.4 \\
JPEG Compression & 17.45 / 98.1 & 17.49 / 98.2 & 14.21 / 98.8 & 12.64 / 99.4 & 9.05 / 99.7  & 14.17 / 98.84 \\
Motion Blur & 21.88 / 98.0 & 17.4 / 98.1 & 14.35 / 98.2 & 12.19 / 98.6 & 10.58 / 98.9  & 15.28 / 98.36 \\
Pixelate & 32.68 / 97.0 & 32.34 / 96.9 & 32.99 / 96.6 & 30.62 / 97.0 & 28.46 / 97.3  & 31.42 / 96.96 \\
Shot Noise & 66.1 / 71.7 & 89.03 / 43.9 & 96.59 / 15.0 & 98.41 / 3.0 & 99.34 / 0.8  & 89.89 / 26.88 \\
Snow & 42.62 / 93.6 & 54.38 / 88.7 & 44.96 / 93.8 & 40.73 / 96.0 & 36.27 / 97.3  & 43.79 / 93.88 \\
Zoom Blur & 16.06 / 98.1 & 12.24 / 98.5 & 10.42 / 98.9 & 8.67 / 99.4 & 7.44 / 99.5  & 10.97 / 98.88 \\
\bottomrule
\end{tabular}}
\caption{The performance of the GLOW model in detecting out-of-distribution (OOD) covariate shift between ImageNet200 and ImageNet200-C datasets is evaluated. The model is evaluated using \textbf{log-likelihood} as metric. The model achieves a mean Area Under the Receiver Operating Characteristic (AUROC) of 36.21\% and a False Positive Rate at 95\% True Positive Rate (FPR95) of 81.7\%.\label{tab:glow_results_ll_tin}}}
\end{table*}

\begin{table*}[htbp]
\centering
{\resizebox{1.0\linewidth}{!}{
\begin{tabular}{r|cccccc}
Severity & 1 & 2 & 3 & 4 & 5 & Average\\
Metric & AUROC$\uparrow$/ FPR95$\downarrow$ & AUROC$\uparrow$/ FPR95$\downarrow$ & AUROC$\uparrow$/ FPR95$\downarrow$ & AUROC$\uparrow$/ FPR95$\downarrow$ & AUROC$\uparrow$/ FPR95$\downarrow$ & AUROC$\uparrow$/ FPR95$\downarrow$  \\
\midrule
Brightness & 43.64 / 96.0 & 37.45 / 97.8 & 33.11 / 98.6 & 29.8 / 99.0 & 28.06 / 99.1  & 34.41 / 98.1 \\
Contrast & 81.03 / 48.7 & 86.93 / 36.3 & 92.61 / 22.5 & 97.37 / 7.0 & 99.17 / 3.4  & 91.42 / 23.58 \\
Defocus Blur & 59.32 / 79.3 & 61.55 / 74.4 & 66.0 / 66.8 & 73.11 / 53.0 & 75.83 / 48.4  & 67.16 / 64.38 \\
Elastic Transform & 56.73 / 85.4 & 57.43 / 83.7 & 59.26 / 79.8 & 58.04 / 80.6 & 56.33 / 83.0  & 57.56 / 82.5 \\
Fog & 65.07 / 73.1 & 71.91 / 62.9 & 76.93 / 54.3 & 82.61 / 45.9 & 86.36 / 36.5  & 76.58 / 54.54 \\
Frost & 44.91 / 95.0 & 46.15 / 93.9 & 48.82 / 91.4 & 49.75 / 90.5 & 50.83 / 90.1  & 48.09 / 92.18 \\
Gaussian Noise & 27.09 / 97.4 & 4.3 / 99.2 & 0.53 / 99.8 & 0.16 / 100.0 & 0.01 / 100.0  & 6.42 / 99.28 \\
Glass Blur & 45.74 / 94.6 & 57.55 / 80.8 & 61.94 / 72.1 & 64.63 / 67.6 & 70.64 / 56.9  & 60.1 / 74.4 \\
Impulse Noise & 22.29 / 98.1 & 4.99 / 99.2 & 0.36 / 100.0 & 0.07 / 100.0 & 0.01 / 100.0  & 5.54 / 99.46 \\
JPEG Compression & 64.51 / 79.8 & 65.75 / 80.9 & 69.9 / 74.4 & 73.9 / 70.8 & 81.06 / 58.7  & 71.02 / 72.92 \\
Motion Blur & 56.63 / 83.9 & 59.15 / 79.1 & 61.21 / 74.4 & 63.11 / 71.6 & 64.74 / 69.4  & 60.97 / 75.68 \\
Pixelate & 52.9 / 90.2 & 53.72 / 89.1 & 53.02 / 89.4 & 55.48 / 84.7 & 60.41 / 78.5  & 55.11 / 86.38 \\
Shot Noise & 27.68 / 97.4 & 13.79 / 99.0 & 5.15 / 99.6 & 2.85 / 99.8 & 1.28 / 100.0  & 10.15 / 99.16 \\
Snow & 51.33 / 92.2 & 47.83 / 94.3 & 54.25 / 91.4 & 57.88 / 90.2 & 63.25 / 85.8  & 54.91 / 90.78 \\
Zoom Blur & 60.25 / 78.2 & 63.37 / 71.6 & 65.24 / 68.4 & 67.39 / 63.8 & 69.11 / 61.2  & 65.07 / 68.64 \\
\bottomrule
\end{tabular}}
\caption{The performance of the GLOW model in detecting out-of-distribution (OOD) covariate shift between ImageNet200 and ImageNet200-C datasets is evaluated. The model is evaluated using \textbf{typicality} as metric. The model achieves a mean Area Under the Receiver Operating Characteristic (AUROC) of 51.0\% and a False Positive Rate at 95\% True Positive Rate (FPR95) of 78.8\%.\label{tab:glow_results_typ_tin}}}
\end{table*}

\begin{table*}[htbp]
\centering
{\resizebox{1.0\linewidth}{!}{
\begin{tabular}{r|cccccc}
Severity & 1 & 2 & 3 & 4 & 5 & Average \\
Metric & AUROC$\uparrow$/ FPR95$\downarrow$ & AUROC$\uparrow$/ FPR95$\downarrow$ & AUROC$\uparrow$/ FPR95$\downarrow$ & AUROC$\uparrow$/ FPR95$\downarrow$ & AUROC$\uparrow$/ FPR95$\downarrow$ & AUROC$\uparrow$/ FPR95$\downarrow$ \\
\midrule
Brightness & 47.03 / 88.7 & 49.81 / 86.4 & 53.01 / 84.6 & 56.25 / 81.5 & 58.86 / 78.5  & 52.99 / 83.94 \\
Contrast & 79.98 / 59.4 & 88.76 / 41.1 & 95.46 / 15.6 & 99.2 / 4.0 & 99.92 / 0.4  & 92.66 / 24.1 \\
Defocus Blur & 55.14 / 76.4 & 59.14 / 70.9 & 65.89 / 61.7 & 76.95 / 46.4 & 80.72 / 41.3  & 67.57 / 59.34 \\
Elastic Transform & 49.51 / 85.9 & 50.52 / 84.5 & 53.73 / 79.7 & 52.89 / 79.5 & 50.35 / 82.4  & 51.4 / 82.4 \\
Fog & 54.47 / 86.2 & 63.48 / 80.7 & 71.4 / 72.8 & 80.79 / 59.6 & 86.43 / 48.1  & 71.31 / 69.48 \\
Frost & 44.2 / 93.3 & 42.28 / 95.6 & 40.22 / 96.2 & 39.61 / 96.2 & 39.12 / 95.6  & 41.09 / 95.38 \\
Gaussian Noise & 48.72 / 94.2 & 90.93 / 17.5 & 97.51 / 4.3 & 99.04 / 1.8 & 99.52 / 0.6  & 87.14 / 23.68 \\
Glass Blur & 47.43 / 95.6 & 47.72 / 87.0 & 56.57 / 75.6 & 62.44 / 66.7 & 73.02 / 50.5  & 57.44 / 75.08 \\
Impulse Noise & 54.18 / 90.5 & 87.27 / 26.8 & 97.76 / 3.9 & 99.29 / 1.4 & 99.77 / 0.4  & 87.65 / 24.6 \\
JPEG Compression & 57.84 / 82.5 & 58.69 / 83.5 & 63.58 / 78.9 & 67.34 / 77.7 & 76.33 / 71.9  & 64.76 / 78.9 \\
Motion Blur & 50.84 / 83.3 & 55.12 / 77.3 & 58.97 / 72.4 & 62.29 / 67.8 & 65.1 / 64.2  & 58.46 / 73.0 \\
Pixelate & 45.1 / 91.9 & 44.29 / 92.1 & 43.55 / 92.0 & 43.3 / 92.0 & 42.39 / 94.2  & 43.73 / 92.44 \\
Shot Noise & 51.85 / 92.6 & 79.07 / 67.4 & 91.32 / 28.1 & 95.42 / 12.3 & 98.25 / 5.0  & 83.18 / 41.08 \\
Snow & 41.77 / 96.6 & 44.85 / 96.7 & 42.98 / 97.0 & 43.99 / 97.1 & 46.22 / 97.0  & 43.96 / 96.88 \\
Zoom Blur & 56.61 / 74.6 & 61.92 / 67.7 & 65.01 / 62.7 & 68.36 / 58.8 & 71.01 / 54.6  & 64.58 / 63.68 \\
\bottomrule
\end{tabular}}
\caption{The performance of the GLOW model in detecting out-of-distribution (OOD) covariate shift between ImageNet200 and ImageNet200-C datasets is evaluated. The model is evaluated using \textbf{normalized score distance} as metric. The model achieves a mean Area Under the Receiver Operating Characteristic (AUROC) of 64.5\% and a False Positive Rate at 95\% True Positive Rate (FPR95) of 65.6\%.\label{tab:glow_results_nsd_tin}}}
\end{table*}

\begin{table*}[htbp]
\centering
{\resizebox{1.0\linewidth}{!}{
\begin{tabular}{r|cccccc}
Severity & 1 & 2 & 3 & 4 & 5 & Average \\
Metric & AUROC$\uparrow$/ FPR95$\downarrow$ & AUROC$\uparrow$/ FPR95$\downarrow$ & AUROC$\uparrow$/ FPR95$\downarrow$ & AUROC$\uparrow$/ FPR95$\downarrow$ & AUROC$\uparrow$/ FPR95$\downarrow$ & AUROC$\uparrow$/ FPR95$\downarrow$ \\
\midrule
Brightness & 24.8 / 98.6 & 32.64 / 97.4 & 40.65 / 96.9 & 44.85 / 96.9 & 50.5 / 96.2  & 38.69 / 97.2 \\
Contrast & 0.86 / 100.0 & 0.35 / 100.0 & 0.12 / 100.0 & 0.03 / 100.0 & 0.02 / 100.0  & 0.28 / 100.0 \\
Defocus Blur & 9.16 / 99.8 & 7.8 / 99.8 & 4.63 / 99.9 & 1.71 / 100.0 & 1.18 / 100.0  & 4.9 / 99.9 \\
Elastic Transform & 12.46 / 99.7 & 12.18 / 99.7 & 9.64 / 99.7 & 9.73 / 99.7 & 10.37 / 99.7  & 10.88 / 99.7 \\
Fog & 6.11 / 99.8 & 2.82 / 100.0 & 1.48 / 100.0 & 0.77 / 100.0 & 0.33 / 100.0  & 2.3 / 99.96 \\
Frost & 23.98 / 98.1 & 22.32 / 97.6 & 17.32 / 98.5 & 16.03 / 98.1 & 15.8 / 97.9  & 19.09 / 98.04 \\
Gaussian Noise & 29.29 / 95.2 & 68.86 / 52.8 & 94.04 / 11.7 & 98.06 / 3.6 & 99.46 / 1.6  & 77.94 / 32.98 \\
Glass Blur & 26.39 / 97.8 & 8.77 / 99.7 & 4.71 / 99.8 & 3.11 / 99.9 & 1.44 / 100.0  & 8.88 / 99.44 \\
Impulse Noise & 42.44 / 88.1 & 71.39 / 60.7 & 96.95 / 6.9 & 99.42 / 1.8 & 99.99 / 0.1  & 82.04 / 31.52 \\
JPEG Compression & 12.31 / 99.5 & 15.34 / 99.5 & 11.5 / 99.7 & 12.1 / 99.7 & 10.84 / 99.8  & 12.42 / 99.64 \\
Motion Blur & 11.93 / 99.7 & 8.59 / 99.7 & 6.98 / 99.8 & 5.11 / 99.8 & 4.29 / 99.9  & 7.38 / 99.78 \\
Pixelate & 17.99 / 99.1 & 16.25 / 99.1 & 15.22 / 99.5 & 14.63 / 99.2 & 12.65 / 99.4  & 15.35 / 99.26 \\
Shot Noise & 29.97 / 95.2 & 52.29 / 82.9 & 80.05 / 50.3 & 92.4 / 20.7 & 98.31 / 4.5  & 70.6 / 50.72 \\
Snow & 23.6 / 97.1 & 30.26 / 94.4 & 28.19 / 95.5 & 26.6 / 96.2 & 23.48 / 97.4  & 26.43 / 96.12 \\
Zoom Blur & 8.49 / 99.8 & 5.74 / 99.8 & 4.61 / 99.8 & 3.86 / 99.8 & 2.8 / 100.0  & 5.1 / 99.84 \\
\bottomrule
\end{tabular}}

\caption{The performance of the CovariateFlow model in detecting out-of-distribution (OOD) covariate shift between ImageNet200 and ImageNet200-C datasets is evaluated. The model is evaluated using \textbf{log-likelihood} as metric. The model achieves a mean Area Under the Receiver Operating Characteristic (AUROC) of 25.8\% and a False Positive Rate at 95\% True Positive Rate (FPR95) of 87.9\%.\label{tab:covariate_results_ll_tin}}}
\end{table*}

\begin{table*}[htbp]
\centering
{\resizebox{1.0\linewidth}{!}{
\begin{tabular}{r|cccccc}
Severity & 1 & 2 & 3 & 4 & 5 & Average\\
Metric & AUROC$\uparrow$/ FPR95$\downarrow$ & AUROC$\uparrow$/ FPR95$\downarrow$ & AUROC$\uparrow$/ FPR95$\downarrow$ & AUROC$\uparrow$/ FPR95$\downarrow$ & AUROC$\uparrow$/ FPR95$\downarrow$ & AUROC$\uparrow$/ FPR95$\downarrow$ \\
\midrule
Brightness & 63.09 / 83.5 & 59.09 / 87.7 & 57.06 / 90.8 & 56.08 / 93.0 & 53.79 / 93.6  & 57.82 / 89.72 \\
Contrast & 80.11 / 52.2 & 82.58 / 46.6 & 84.46 / 43.7 & 86.7 / 38.5 & 87.91 / 38.4  & 84.35 / 43.88 \\
Defocus Blur & 73.57 / 65.0 & 76.69 / 59.2 & 77.32 / 56.7 & 79.52 / 52.1 & 80.14 / 51.6  & 77.45 / 56.92 \\
Elastic Transform & 70.95 / 76.5 & 71.3 / 69.2 & 71.14 / 68.0 & 71.97 / 66.7 & 69.81 / 68.8  & 71.03 / 69.84 \\
Fog & 73.8 / 64.2 & 77.47 / 56.5 & 78.13 / 54.9 & 79.93 / 50.5 & 81.97 / 46.8  & 78.26 / 54.58 \\
Frost & 62.49 / 81.5 & 61.63 / 81.4 & 65.72 / 76.1 & 64.46 / 76.5 & 66.97 / 72.0  & 64.25 / 77.5 \\
Gaussian Noise & 57.59 / 86.2 & 37.25 / 95.9 & 11.56 / 99.4 & 5.28 / 99.9 & 2.63 / 100.0  & 22.86 / 96.28 \\
Glass Blur & 57.5 / 84.7 & 70.68 / 69.2 & 73.88 / 62.4 & 75.91 / 58.8 & 79.3 / 52.9  & 71.45 / 65.6 \\
Impulse Noise & 50.98 / 92.7 & 37.38 / 96.1 & 9.52 / 99.7 & 2.75 / 100.0 & 0.58 / 100.0  & 20.24 / 97.7 \\
JPEG Compression & 68.64 / 72.4 & 66.91 / 75.4 & 70.38 / 70.6 & 69.92 / 74.2 & 69.95 / 71.5  & 69.16 / 72.82 \\
Motion Blur & 69.2 / 70.4 & 73.33 / 65.6 & 74.94 / 62.2 & 76.46 / 59.5 & 76.41 / 60.0  & 74.07 / 63.54 \\
Pixelate & 65.93 / 77.7 & 68.28 / 75.3 & 70.48 / 68.9 & 71.53 / 68.9 & 72.87 / 66.4  & 69.82 / 71.44 \\
Shot Noise & 56.38 / 83.8 & 46.32 / 93.2 & 27.58 / 97.8 & 15.13 / 99.4 & 5.62 / 100.0  & 30.21 / 94.84 \\
Snow & 60.07 / 81.9 & 56.09 / 84.8 & 57.87 / 83.5 & 59.16 / 83.5 & 62.24 / 79.2  & 59.09 / 82.58 \\
Zoom Blur & 74.45 / 61.1 & 75.93 / 61.7 & 75.98 / 59.3 & 76.89 / 59.0 & 77.85 / 57.1  & 76.22 / 59.64 \\
\bottomrule
\end{tabular}}

\caption{The performance of the CovariateFlow model in detecting out-of-distribution (OOD) covariate shift between ImageNet200 and ImageNet200-C datasets is evaluated. The model is evaluated using \textbf{typicality} as metric. The model achieves a mean Area Under the Receiver Operating Characteristic (AUROC) of 61.8\% and a False Positive Rate at 95\% True Positive Rate (FPR95) of 73.1\%.\label{tab:covariate_results_typicality_tin}}}
\end{table*}

\begin{table*}[htbp]
\centering
{\resizebox{1.0\linewidth}{!}{
\begin{tabular}{r|ccccccc}
Severity & 1 & 2 & 3 & 4 & 5 & Average\\
Metric & AUROC$\uparrow$/ FPR95$\downarrow$ & AUROC$\uparrow$/ FPR95$\downarrow$ & AUROC$\uparrow$/ FPR95$\downarrow$ & AUROC$\uparrow$/ FPR95$\downarrow$ & AUROC$\uparrow$/ FPR95$\downarrow$ & AUROC$\uparrow$/ FPR95$\downarrow$  \\
\midrule
Brightness & 58.65 / 86.1 & 54.19 / 93.5 & 53.38 / 93.2 & 53.62 / 94.2 & 54.55 / 94.1  & 54.88 / 92.22 \\
Contrast & 87.87 / 23.7 & 89.66 / 19.9 & 91.65 / 16.4 & 93.19 / 14.0 & 94.38 / 12.0  & 91.35 / 17.2 \\
Defocus Blur & 75.17 / 69.5 & 76.79 / 63.5 & 80.82 / 48.0 & 85.66 / 28.3 & 87.26 / 24.3  & 81.14 / 46.72 \\
Elastic Transform & 71.4 / 71.1 & 70.95 / 76.3 & 72.68 / 72.8 & 73.58 / 71.4 & 71.72 / 72.8  & 72.07 / 72.88 \\
Fog & 78.02 / 52.0 & 83.44 / 36.2 & 85.75 / 27.8 & 87.64 / 23.2 & 89.49 / 19.3  & 84.87 / 31.7 \\
Frost & 57.73 / 87.8 & 57.58 / 87.6 & 61.2 / 82.5 & 63.14 / 76.3 & 63.24 / 78.7  & 60.58 / 82.58 \\
Gaussian Noise & 51.71 / 91.5 & 42.24 / 94.7 & 89.85 / 18.7 & 95.72 / 7.2 & 97.91 / 3.6  & 75.49 / 43.14 \\
Glass Blur & 54.51 / 89.5 & 73.88 / 67.8 & 79.7 / 46.0 & 82.48 / 36.7 & 86.13 / 26.8  & 75.34 / 53.36 \\
Impulse Noise & 44.24 / 94.0 & 49.66 / 94.9 & 94.03 / 12.2 & 97.96 / 3.9 & 99.33 / 1.0  & 77.04 / 41.2 \\
JPEG Compression & 69.16 / 73.0 & 66.41 / 82.2 & 69.87 / 75.2 & 69.76 / 74.7 & 71.05 / 75.4  & 69.25 / 76.1 \\
Motion Blur & 70.49 / 77.0 & 74.58 / 66.4 & 76.85 / 63.0 & 79.64 / 53.8 & 80.66 / 45.2  & 76.44 / 61.08 \\
Pixelate & 64.6 / 82.7 & 65.71 / 81.4 & 65.98 / 81.7 & 67.72 / 83.3 & 68.87 / 77.1  & 66.58 / 81.24 \\
Shot Noise & 49.73 / 90.8 & 37.74 / 96.7 & 66.7 / 83.8 & 87.36 / 36.0 & 96.12 / 9.1  & 67.53 / 63.28 \\
Snow & 55.22 / 88.2 & 48.07 / 90.9 & 50.33 / 92.4 & 51.18 / 91.8 & 54.07 / 90.6  & 51.77 / 90.78 \\
Zoom Blur & 74.91 / 68.0 & 79.54 / 49.0 & 80.36 / 46.6 & 81.94 / 42.2 & 83.66 / 34.6  & 80.08 / 48.08 \\
\bottomrule
\end{tabular}}

\caption{The performance of the CovariateFlow model in detecting out-of-distribution (OOD) covariate shift between ImageNet200 and ImageNet200-C datasets is evaluated. The model is evaluated using \textbf{normalized score distance} as metric. The model achieves a mean Area Under the Receiver Operating Characteristic (AUROC) of 72.3\% and a False Positive Rate at 95\% True Positive Rate (FPR95) of 60.1\%.\label{tab:covariate_results_nsd_tin}}}
\end{table*}

\newpage
\subsection{Ablation Study}\label{sec:supp_ablation}
\begin{figure}[htbp]
 \begin{center}
  \includegraphics[width=0.95\linewidth]{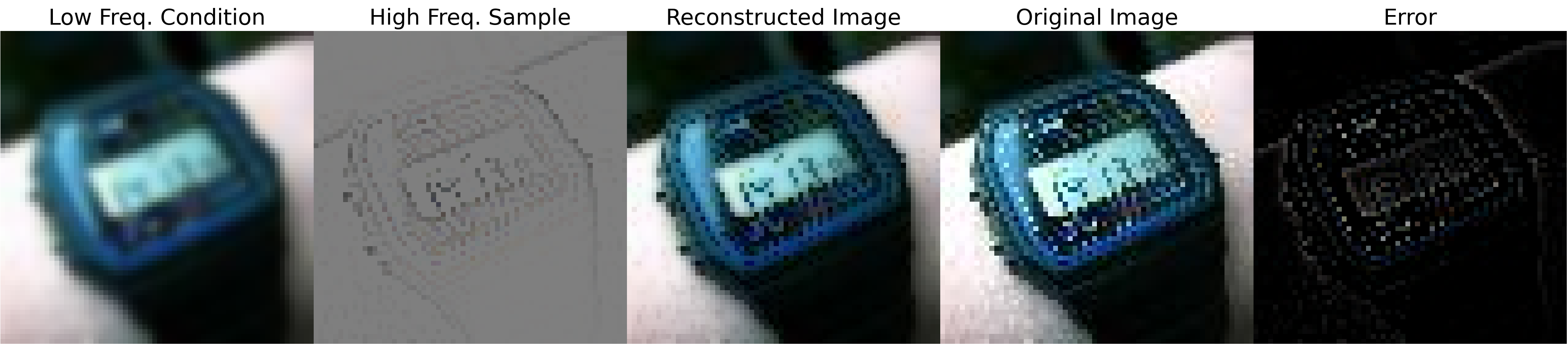}
  \caption{Example plausible high-frequency sample when conditioned on a low frequency image. The sample clearly shows the conditioning on the low-frequency image, with high-frequency components generated along the watch edges, the time and more uniformly distributed background noise on the persons' arm. }
  \label{fig:sample}
 \end{center}
\end{figure}

This section details a series of ablation experiments conducted, including an analysis of the effect of the individual components in CovariateFlow on the detection performance (Table~\ref{tab:ablation_result}), mean scores per severity of the models and resource aspects are depicted in Table~\ref{tab:model_details}, model performance on a typically semantic OOD detection problem in Table~\ref{tab:semantic_OOD} and finally an example (Figure~\ref{fig:sample}) of heteroscedastic high-frequency components sampled from the fully invertible CovariateFlow.

In our ablation experiments, we test the effect of explicitly modelling the conditional distribution between the low-frequency and high-frequency signal components as described in Section~\ref{sec:covariate_flow}. This is achieved by training and evaluating the CovariateFlow model in four different settings: (1)~unconditional coupling flows with the full input image, (2)~unconditional coupling flows subject to only the high-frequency components of the image, (3)~unconditional coupling flows subject to the high-frequency components and a conditional signal-dependent layer additionally subject to the low-frequency image components and finally, (4)~the high-frequency image components applied to the conditional coupling flows and a signal dependent layer subject to the low-frequency components. For each of these implementations we follow the exact same training methodologies as described in Section~\ref{sec:supp_implemntation_details}. All the images are encoded at 16 bit depth during dequantization to ensure comparability. 

\begin{table}[!htbp]
\centering
{\resizebox{1.0\linewidth}{!}{
\begin{tabular}{rcc} 
\toprule
    &        \underline{CIFAR10}        &  \underline{CIFAR10-C} \\
Model \enspace &\enspace mean BPD$\downarrow$ \enspace &\enspace LL$\uparrow$/ Typicality$\uparrow$/ NSD$\uparrow$ \\
\midrule
Full Image Unconditional (1) \enspace &\enspace 11.32 & 56.6 / 64.2 / 67.8 \\
High Frequency \& Unconditional (2) \enspace &\enspace 9.85 & 57.9 / 40.8 / 65.5  \\
High Frequency \& Unconditional + SDL (3) \enspace &\enspace 9.77 & 58.4 / 40.0 / 62.6  \\
High Frequency \& Conditional + SDL (4) \enspace &\enspace \textbf{5.48} & 58.3 / 45.5 / \textbf{74.9} \\
\bottomrule 
\end{tabular}}
\caption{Results (Bits per dimension or AUROC) obtained from the ablation Experiments on CovariateFlow with CIFAR10(-C).\label{tab:ablation_result}}}
\end{table}
From Table~\ref{tab:ablation_result} it can be seen that while model 1 is limited in modelling the complete data distribution (11.32 Bits per dimension (BPD)), it performs well on detecting OOD covariate shift with NSD (AUROC 67.8\%), comparable to the performance obtained with GLOW. Only using the high-frequency image components in an unconditional setting (model 2) yields a somewhat lower OOD detection performance of 65.5\% AUROC. Introducing the SDL (model 3), lowers the mean BPD and improves on LL-based OOD detection (58.4\%), but adversely effects the NSD evaluation (62.6\%). While the SDL layer does not show improvement in the detection performance, it significantly aided in satabalizing model training. Finally (model 4), conditioning every coupling flow in the network on the low-frequency content significantly improves modelling the high-frequency components (9.77 BPD $\rightarrow$ 5.48 BPD), indicating the value in the additional information. Modelling this conditional relation between the low-frequency and high-frequency components also proves very effective in detecting OOD covariate shift. The model achieves a mean AUROC of 74.9\% at detecting covariate factors across all variations and degredations when evaluated with NSD. 
\begin{table}[!htbp]
\centering
{\resizebox{1.0\linewidth}{!}{
\begin{tabular}{rcccccccc} 
\toprule
 & \multicolumn{5}{c}{Mean distance / CIFAR10-C severity} & Models Size & Inference Speed \\
Model & 1& 2&3&4&5 & (\# parameters) & milliseconds\\
\midrule
Vanilla VAE~\cite{kingma2022autoencoding} (SSIM + KL Div) & \enspace 0.0365 \enspace & \enspace 0.0367 \enspace & \enspace 0.0381 \enspace & \enspace 0.0408 \enspace & \enspace 0.0430 \enspace & 9,436,867 & 4.1ms\\
AVAE~\cite{plumerault2021avae} (MSE + KL Div + Adv Loss) & 0.066& 0.073& 0.078&0.086&0.0992 & 11,002,373 & 9.1ms\\
DDPM~\cite{graham2023denoising} (T20: LPIPS)& 0.7&0.8&0.9&1.0&1.4 & 17,714,563 & 34.1ms\\
DDPM~\cite{graham2023denoising} (T20: LPIPS + MSE)& 1.6&1.8&2.3&2.8&3.6 & 17,714,563 & 34.1ms\\
GLOW~\cite{kingma2018glow} (LL) & 0.73 & 1.1& 1.2&1.4 &238,10.9& 44,235,312& 65.8ms\\
GLOW~\cite{chali2023typicality} (Typicality) & 2.2&2.8&2.8&2.9&3.2 & 44,235,312&178.3ms \\
GLOW (NSD) & 1.2 & 1.7 & 1.9 & 2.2 & 411,753.0 & 44,235,312& 178.3ms\\
CovariateFlow (LL) & 1.1 &1.5&1.7&2.0&2.2 & 945,882& 22.5ms \\
CovariateFlow (Typicality) & 0.02&0.03&0.04&0.05&0.07 & 945,882& 59.6ms\\
CovariateFlow (NSD) & 2.3&2.9&3.4&3.7&4.3 & 945,882& 59.6ms\\
\bottomrule 
\end{tabular}}}
\caption{Model specific details and results. The mean distance (measured differently per model) per severity, the number of trainable parameters and the inference time are depicted. Note that the DDPM is evaluated multiple times to obtain a detection score.}
\label{tab:model_details}
\end{table}
Table~\ref{tab:model_details} presents additional information about each of the models employed in this research. This table showcases the mean distance measurements (CIFAR10-C), taken under different evaluation criteria, across increasing severity levels of covariate shifts within the dataset. Such a detailed breakdown allows for a nuanced understanding of each model's resilience and adaptability to changes in input data distribution. Notably, the LL evaluations of GLOW at the highest severity level encountered numerical stability issues, leading to the substitution of some results with the maximum representable floating-point number. This adjustment, while necessary, underscores the challenges in maintaining computational integrity under extreme conditions and the importance of implementing robust handling mechanisms for such anomalies. It is evident from the data that there is a consistent trend of increasing mean distance scores across all models as the severity level escalates, highlighting the impact of covariate shift on model performance. This trend underscores the ability to quantify covariate shift, although only briefly evaluated here. Furthermore, the table delineates the model size, quantified by the number of trainable parameters, and the inference speed, measured in milliseconds. These metrics are critical for understanding the trade-offs between model complexity, computational efficiency, and performance.The data presented in Table~\ref{tab:model_details} not only elucidates the ability to quantify covariate shifts, but also emphasizes the importance of balancing model complexity and computational efficiency when considering the model deployment conditions. 

\begin{wraptable}{r}{0.45\textwidth}
\centering
\vspace{-0.5cm}
  {\resizebox{1.0\linewidth}{!}{
  
  \begin{tabular}{cl|c}
  & & OOD SVHN~\cite{svhn}\\
  & Models & AUROC $\uparrow$ \\
  \cmidrule{1-3}
  \parbox[t]{5mm}{\multirow{12}{*}{\rotatebox[origin=c]{90}{CIFAR10 ID}}} & \textbf{Reconstruction}\\
  \cmidrule{2-3}
   &DDPM~\cite{graham2023denoising} & \hspace*{1mm}\textbf{97.9*}/ 95.8 \\
   \cmidrule{2-3}
   &\textbf{Explicit Density}\\
   \cmidrule{2-3}
   &Vanilla VAE (SSIM + KL Div) & 24.4\\
   &AVAE (MSE + KL Div + Adv Loss) \hspace*{3mm} & 32.0\\
   &VAE-FRL~\cite{cai2023out} & 85.4* \\
   &GLOW-FRL~\cite{cai2023out} & 91.5* \\
   &GLOW (LL) & 0.7\\
   &GLOW (Typicality) & 91.3\\
   &GLOW (NSD) & 89.9\\
   &CovariateFlow (LL) & 0.3\\
   &CovariateFlow (Typicality) & 89.9\\
   &CovariateFlow (NSD) & 90.0\\
   \bottomrule
  \end{tabular}}}
  \caption{The performance of various models on detecting SVHN as OOD when trained on CIFAR10 as ID. * indicates values taken from the published paper.}
  \label{tab:semantic_OOD}
  \vspace{-0.5cm}
\end{wraptable}

Modeling the conditional distribution between the low-frequency and high-frequency components using CovariateFlow is highly effective in detecting out-of-distribution (OOD) covariate shifts. The CIFAR10 dataset, known for its diversity, encompasses a range of in-distribution (ID) covariate conditions. When assessing CovariateFlow in the context of a semantic OOD detection problem, such as distinguishing between CIFAR10 and SVHN datasets, it is plausible that some covariate conditions in CIFAR10 overlap with those in the SVHN dataset. Despite this potential overlap, CovariateFlow demonstrates robust performance in identifying the OOD covariate conditions present in the SVHN dataset, as evidenced by the results shown in Table~\ref{tab:semantic_OOD}.  Although the DDPM (utilizing all 1000 starting points) achieves the best performance, CovariateFlow offers competitive results. This is notable given its significantly smaller size and its specific design focus on covariate conditions rather than semantic content. 

\end{document}